\documentclass[runningheads]{llncs}

\usepackage[review,year=2026,ID=7580]{eccv}

\let\maketitle\maketitleold

\usepackage{eccvabbrv}
\usepackage{graphicx}
\usepackage{booktabs}
\usepackage[accsupp]{axessibility}
\usepackage{float}

\usepackage[breaklinks,colorlinks,citecolor=eccvblue]{hyperref}
\usepackage{orcidlink}

\usepackage{makecell}
\usepackage{booktabs} 

\usepackage[table]{xcolor}  
\definecolor{rowblue}{RGB}{230,244,255}
\definecolor{rowgreen}{RGB}{232,246,232}

\newcommand{\NImg}{\ensuremath{N}}                     

\newcommand{\Feat}{\ensuremath{d}}     

\usepackage{multirow}

\begin{document}

\nolinenumbers

\title{Lightweight Neural Framework for Robust 3D Volume and Surface
Estimation from Multi-View Images}

\author{
Diego E. Farchione \and
Ramzi Idoughi \and
Peter Wonka
}

\institute{
King Abdullah University of Science and Technology (KAUST), Saudi Arabia\\
\email{\{diegoeustachio.farchione,ramzi.idoughi,peter.wonka\}@kaust.edu.sa}
}

\maketitle
\markboth{}{}

\begin{abstract}
Accurate volume and surface area estimation is critical for diverse applications, from marine ecology to medical diagnostics. However, existing methods often suffer from high computational costs and poor performance with sparse and noisy data. We propose a fully feed-forward framework that regresses scale-normalized volume and surface area and their associated uncertainties directly from multi-view images. By fusing 3D point cloud reconstructions with view-aligned 2D features through a graph-based decoder, our model bypasses iterative optimization, ensuring exceptional scalability and rapid inference. Experimental results demonstrate that our approach outperforms state-of-the-art methods, particularly when operating with a low number of input images. Validated across coral monitoring, dietary analysis, and anthropometry, our proposed framework provides a robust, adaptable solution for quantitative shape analysis. This architecture provides a high-speed, scalable alternative for precise geometric estimation from visual data, maintaining high performance even in resource-constrained or sparse-view scenarios.
\end{abstract}

\begin{figure}[H]
  \centering
  \includegraphics[width=0.65\textwidth]{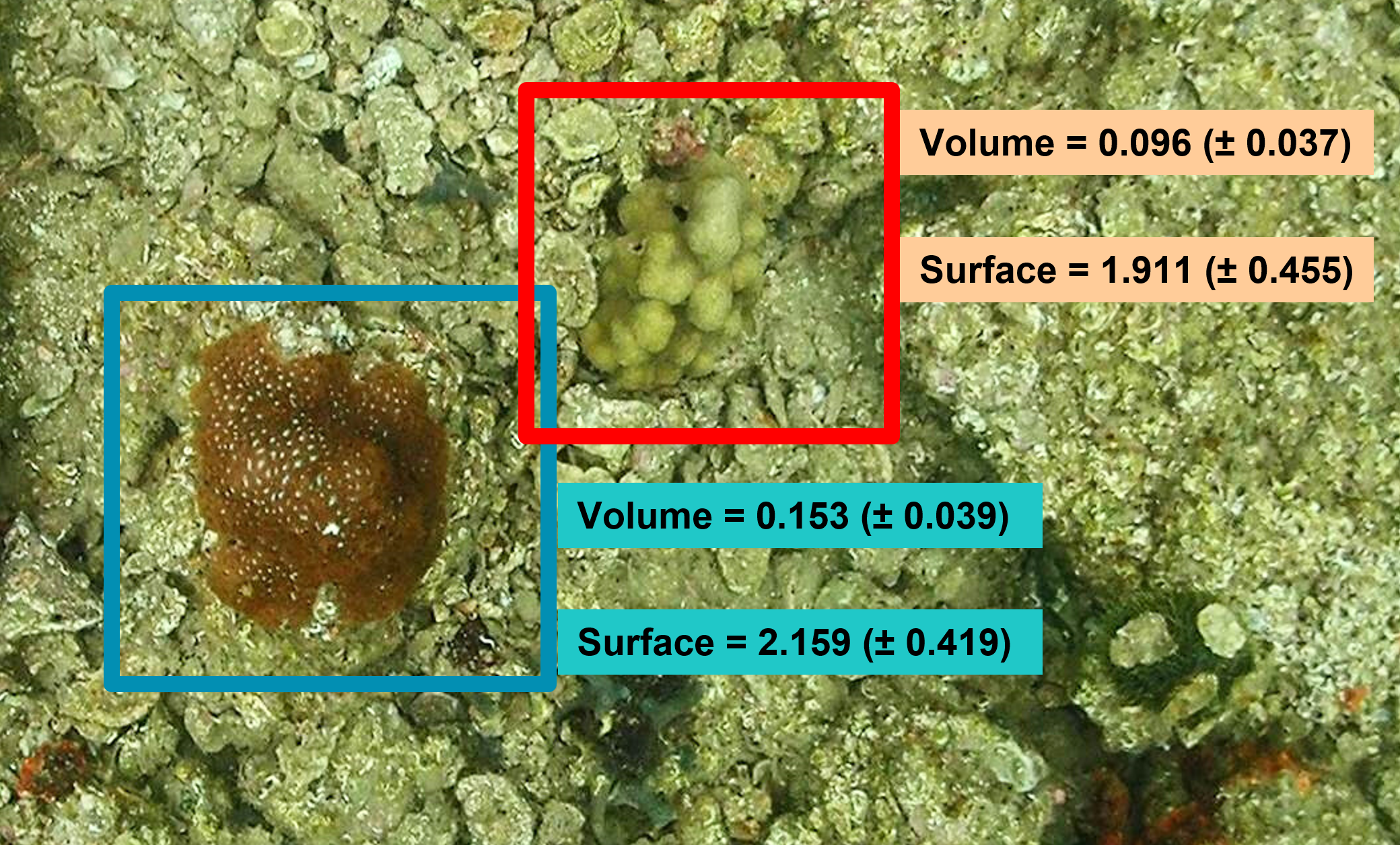}
\caption{\textbf{Teaser}: Our end-to-end pipeline for robust volume and surface area estimation from multi-view images is demonstrated on corals from the CoralVOS dataset~\cite{ziqiang2023coralvos}: Using only five top-view RGB images from monocular video, our model can predict normalized volume and surface with their corresponding uncertainty, showcasing its potential for efficient, real-world applications.}
\label{fig:teaser}
\end{figure}

\section{Introduction}
\label{sec:intro}
Estimating the volume and surface area of objects from visual data is a critical task with profound implications across a wide range of domains. These measurements enable different applications, such as coral growth monitoring, which is a key element in reef conservation efforts, personalized dietary planning, and anthropometry for medical diagnostics. Yet, this task remains challenging due to the inherent complexities of deriving accurate 3D properties from 2D images. Conventional techniques typically rely on manual measurements or explicit 3D mesh generation. For coral monitoring, traditional pipelines~\cite{yudelman2022coral,helmholz2024evaluating} often require specialized scanning, human supervision, and heavy post-processing, limiting scalability for large-scale surveys. Similar issues arise across domains: sparse views, watertight-mesh requirements, calibration errors, and reconstruction artifacts reduce accuracy and increase latency.

In recent years, the evolution of deep learning has transformed visual data processing, offering promising avenues to overcome these challenges. On one hand, General-purpose 3D reconstruction systems like the Visual Geometry Grounded Transformer (VGGT)~\cite{wang2025vggt} and MapAnything~\cite{keetha2025mapanything} enable rapid point cloud generation from multi-view inputs with far less hand-tuning than classical pipelines. Meanwhile, point-cloud networks, such as PointNet/PointNet++~\cite{qi2017pointnet, qi2017pointnetplusplus}, DGCNN~\cite{wang2019dynamic}, PointTransformer~\cite{zhao2021point}, GCN~\cite{kipf2017semi}, and GIN~\cite{xu2019how}, excel at learning directly from unordered sets with locality and invariance priors. On the other hand, state-of-the-art 2D encoders such as DINOv3~\cite{simeoni2025dinov3} provide robust feature extraction from images. Together, these developments facilitate end-to-end frameworks that integrate multimodal information by combining the strengths of 2D visual cues with 3D structural insights to enable more efficient quantitative metric predictions. However, most existing systems still focus on intermediate tasks like geometry reconstruction or image synthesis, relegating volume and surface estimation to post-processing, thereby limiting their practicality and generalizability across diverse scenarios. 

To address these gaps, we propose a \emph{Lightweight Neural Framework for Robust 3D Volume and Surface Estimation from Multi-View Images}. Because absolute scale is ambiguous in unconstrained RGB multi-view capture, we predict \emph{scale-normalized} volume and surface area under a unit-extent normalization. This feed-forward approach begins by masking the object of interest using advanced promptable segmentation tools for precise isolation or collecting synthetic images of the object. Next, it applies general-purpose multi-view reconstructors to generate a detailed point cloud from the processed views. A compact point-cloud network then serves as an decoder to derive a 3D latent vector from this representation. Simultaneously, features are extracted from the 2D images using a state-of-the-art encoder like DINOv3, and are merged with the 3D latent vector. Finally, a Fully Connected (FC) network estimates scale-normalized volume and surface metrics along with their confidence levels, achieved through two parallel branches that employ specialized encoder-decoder networks for each metric. 
Unlike prior estimation methods that rely on explicit meshing and/or iterative optimization, we show that a lightweight 2D–3D fusion regressor trained with evidential uncertainty can directly predict scale-normalized volume and surface area from sparse, noisy multi-view reconstructions across multiple domains.

The key contributions of our approach are as follows:

\begin{itemize}
    \item \textbf{Streamlined end-to-end metrics regression.} A lightweight and computationally efficient framework that directly predicts volume and surface area from masked multi-view inputs, bypassing traditional mesh generation and post-processing for greater efficiency. Thus, our framework supports efficient deployment across large-scale applications without compromising performance.
    \item \textbf{Integrated 2D-3D feature fusion.} By combining advanced 2D feature extraction with compact point-cloud encoding and parallel prediction branches, our method enhances robustness to challenges like sparse views and noise.
    \item \textbf{Deep evidential uncertainty estimation.} Our framework employs a Deep Evidential Regression objective to provide calibrated confidence estimates for volume and surface measurements. By utilizing a Normal-Inverse-Gamma prior and optimizing a composite marginalized likelihood and regularization loss, the model explicitly quantifies aleatoric and epistemic uncertainties,  establishing greater reliability for scientific applications.
    \item \textbf{Extensive evaluation on synthetic and real-world data.} Through comprehensive assessments on both synthetic and real datasets, we showcase superior accuracy, robustness, particularly in handling imperfect data scenarios, and effective generalization to various fields, such as coral monitoring, dietary planning, and anthropometry.
\end{itemize}

\begin{figure*}[t]
  \centering
  \includegraphics[width=0.92\textwidth]{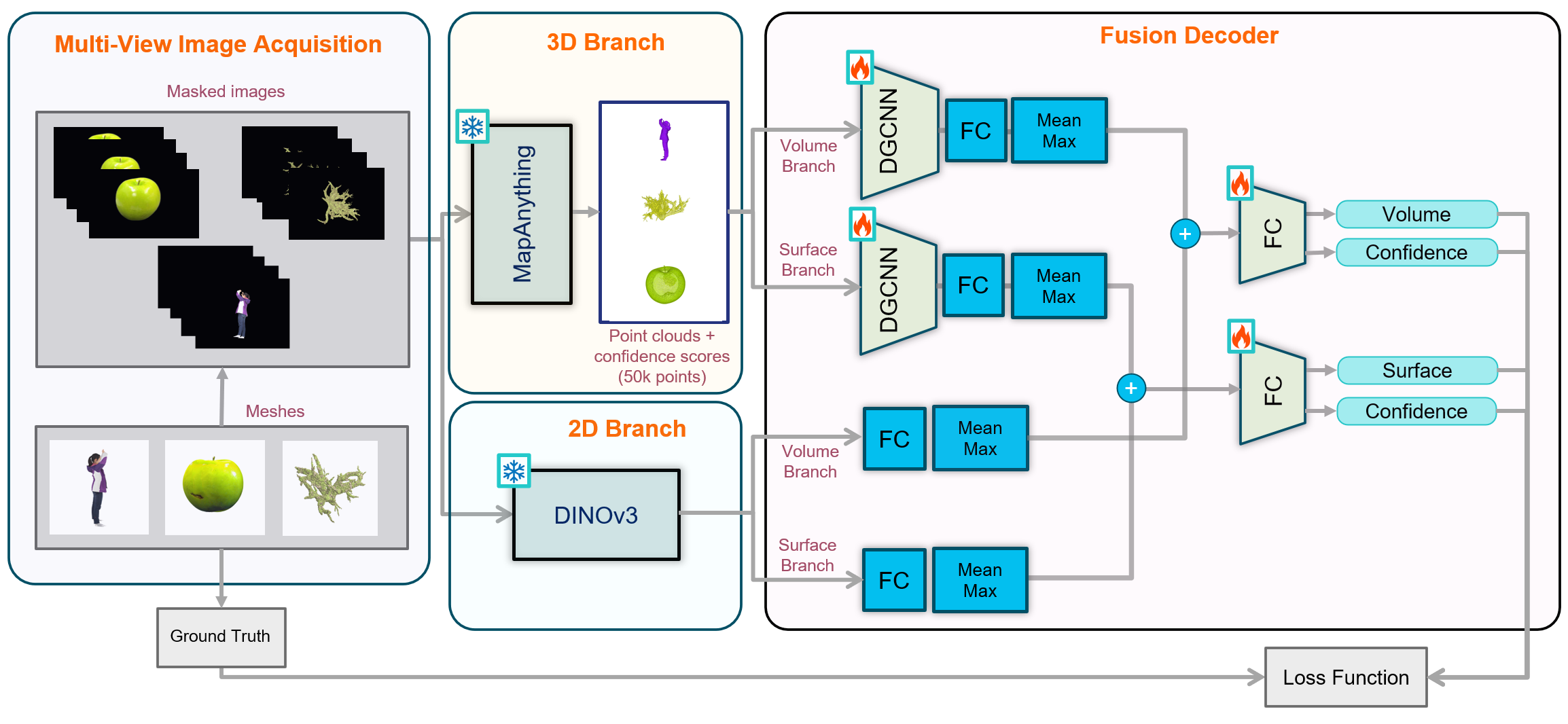}
  \caption{From masked multi-view images from meshes (or original images), a 3D reconstructor like MapAnything produces a fused point cloud with per-point confidence, while a frozen encoder extracts view-aligned 2D features. A lightweight decoder summarizes the point cloud, followed by max pooling layer. In the 2D branch, per-view features pass through a small FC block and are mean-max pooled across views. The fusion decoder concatenates the 3D and 2D descriptors for each target and feeds a small FC regressor that outputs Volume and Surface along with their Epistemic and Aleatoric uncertainties. A composite loss (Evidential loss + MAPE + L1) trains the heads end-to-end, yielding single-pass volume/surface estimates without the need to reconstruct watertight meshes or applying heavy post-processing.}
  \label{fig:pipeline}
\end{figure*}

\section{Related Work}
\label{sec:related}

\subsection{3D Reconstruction from Multi-View Images}

Traditional multi-view stereo (MVS) pipelines, such as those based on Structure from Motion (SfM) and dense reconstruction (e.g., COLMAP~\cite{schonberger2016structure} or PMVS~\cite{furukawa2010accurate}), rely on photometric consistency and keypoint matching (e.g., SIFT~\cite{lowe2004distinctive}) to generate point clouds or meshes. While effective, these methods often require dense views, accurate calibration, and extensive post-processing, making them computationally intensive and prone to failures in textureless or sparse-view scenarios.

Recent neural approaches have improved efficiency and robustness. Methods like NeRF~\cite{mildenhall2020nerf} and its variants (e.g., Mip-NeRF~\cite{barron2021mip}, Instant-NGP~\cite{mueller2022instant}) enable implicit 3D scene reconstruction from multi-view images, while explicit techniques like 3D Gaussian Splatting~\cite{kerbl20233d} accelerate rendering. Concurrently, large-scale 3D foundation models like Hunyuan~\cite{zhao2025hunyuan3d20scalingdiffusion} and Trellis~\cite{xiang2025trellis} have emerged, leveraging powerful generative priors to synthesize high-quality 3D assets directly from single or sparse images. Alongside these generative approaches, generalizable models such as VGGT~\cite{wang2025vggt} and MapAnything~\cite{keetha2025mapanything} extend scene reconstruction to sparse inputs by learning view-invariant features. However, these systems typically focus on geometry recovery and require separate post-processing for volume and surface area estimation, limiting their applicability in real-time or noisy settings. Our framework leverages these advancements (e.g., point cloud generation) but extends them by integrating 2D features and directly regressing metrics, thereby avoiding the need for heavy mesh-based workflows.

\subsection{Volume and Surface Estimation in Specific Domains}
In domain-specific applications, volume estimation has been extensively explored. For food items, early methods used multi-view stereo~\cite{dehais2017two} or RGB-D sensors~\cite{gonzalez2024preliminary, shao2023vision} to reconstruct 3D models of the food for dietary analysis. Other works leverage single RGB images, often relying on fiducial markers to resolve scale ambiguity~\cite{yue2010, fang2016comparison, naritomi2021hungry, Vasiloglou2021, kadam2022fvestimator}. Recent deep learning approaches, such as VolEta~\cite{almughrabi2024voleta} and VolTex~\cite{almughrabi2025voltex}, bypass 3D reconstruction to directly regress volume. While improving scalability, these methods still frequently assume controlled environments (e.g., checkerboard patterns~\cite{vinod2024food}) and struggle with sparse views or occlusions.

Estimating human-body volume is important for monitoring obesity, muscle atrophy, and lymphedema. Most methods rely on structured-light/RGB-D scans to reconstruct meshes and compute volume, requiring subjects to remain still and often involving error-prone post-processing such as surface completion, even from partial depth~\cite{lunscher2018deep}. To avoid this, more recent models bypass full reconstruction altogether, using single RGB or depth images to directly regress part-specific volumes (e.g., VolNet~\cite{leinen2021volnet}, Point2PartVolume~\cite{hu2023point2partvolume}). These methods commonly draw on parametric body models, like SMPL~\cite{loper2023smpl}, to enforce plausible human shapes.


Similarly, ecological applications like coral monitoring~\cite{koch20213d, lange2020quick, yudelman2022coral, helmholz2024evaluating} rely heavily on SfM or tomography, demanding high-fidelity inputs and substantial manual oversight. Across all these domains, the persistent reliance on dense data, controlled setups, and extensive mesh post-processing underscores the critical demand for the robust, end-to-end regression framework we propose.

Beyond volume, surface area estimation has seen specialized development. While some learning-based methods, such as those by Liu et al. ~\cite{liu2022learning}, focus specifically on predicting the visible surface area of irregular objects, most literature treats surface area as a secondary metric derived from 3D scans. In these cases, researchers typically utilize analytical or algorithmic measurements on reconstructed meshes, a methodology extensively applied in body surface area estimation from 3D scans ~\cite{ng2016clinical}. Similar pipelines are prevalent in medical imaging, where 3D reconstructions, often driven by robotic systems or high-resolution scanners, are post-processed to segment and measure wounds ~\cite{filko20232d, liu2019wound} or skin lesions ~\cite{mirzaalian2019measuring}.

\section{Methods}
\label{sec:methods}

\subsection{Overview}
Our pipeline architecture, as shown in Figure~\ref{fig:pipeline}, takes as input a set of multi-view RGB images of a single object and is trained to output both the volume and surface area, along with their associated confidence. It consists of four main components: (1) This pre-processing step segments the target object to generate masked multi-view images, isolating the object from its background. This segmentation step is performed using Grounded-SAM2, which combines Grounding DINO and SAM~2 for promptable, open-world segmentation~\cite{ren2024grounded,liu2023grounding,kirillov2023segany,ravi2024sam2segmentimages,ren2024grounding,jiang2024trex2} or simply by generating images of the object.  (2) \textbf{3D Branch:} This component recovers and fuses per-view geometry from the masked images into a unified point cloud with associated confidences, which serves as the geometric backbone after a simple normalization. (3) \textbf{2D Branch:} The third component extracts 2D features from the masked views. These components are based on foundational models, which are not fine-tuned as part of our pipeline. (4) \textbf{Fusion decoder:} In this final step, two parallel architectures independently regress the volume and surface area. Each architecture employs a GIN-style network augmented with a shallow fully connected layer to encode the 3D point cloud into a latent code; on the 2D side, a shallow fully connected network with a pooling layer encodes the multi-view 2D-encoded features into a compact global descriptor. Regression is then performed by an FC-based decoding head on the concatenated latent code and global descriptor. Uncertainties are also predicted using an uncertainty-aware loss. In the following, we describe each component of our framework in detail.

\subsection{Features Extraction}
\label{subsec:features}
To enable accurate estimation of volume and surface area for 3D objects, we combine complementary 2D and 3D representations derived from sparse multi-view inputs. Indeed, the 3D representation (e.g., point cloud) captures essential geometric shape and spatial dimensions, while the 2D image conveys visual details like textures and edges. This integration ensures a more robust understanding by leveraging both modalities, preventing the limitations of relying on either alone.

\noindent \textbf{2D Feature Extraction:} We utilize a DINOv3 ViT-S/16 encoder (with a patch size of 16 and hidden/pooler dimensionality of \(\mathbf{384}\)) to process each masked RGB view and generate embedded features. This approach extracts global, view-aggregated descriptors that highlight visual characteristics. 

\noindent \textbf{3D Feature Extraction:} For the 3D representation, we employ a general-purpose multi-view reconstruction model to produce a sparse, fused point cloud with per-point confidence scores. In our primary setup, we leverage the MapAnything framework with optimized settings for efficiency, applied to masked inputs. The model outputs 3D points and confidence values per view, which we aggregate across views and augment with confidence as an additional channel, resulting in features of the form $\big( x, y, z, conf\big )$. We post-process this feature vector using two filters: one removes points mapped to background pixels, and the other suppresses low-confidence artifacts, producing a cleaner point cloud. 

\subsection{Feature Fusion and Regression}
In the final module, we use two independent heads (same architecture) to regress volume and surface area. We fuse 3D point-cloud features and multi-view 2D features into a unified representation, enabling accurate estimation of the object's properties. The 3D features, derived from the point cloud, are processed through a graph-based network that aggregates point-wise information into a compact embedding, ensuring permutation invariance, which is further refined using max pooling and a multi-layer perceptron (MLP) to produce the final 3D latent code, denoted as: $z_{3D}$. Meanwhile, the 2D features from multiple views are aggregated into a single global descriptor $\Feat_{2D}$, achieved by passing each view's features through a shallow MLP and then combining them via mean and max pooling across views.
At this stage, we concatenate the 2D and 3D representation features $\big[\, z_{3D} \,\|\, \Feat_{2D} \,\big]$ and feed it to a last FC network that acts as the regression head and outputs. Rather than outputting a simple point estimate and variance, the regression head outputs four distinct parameters ($\gamma, v, \alpha, \beta$) that define a Normal-Inverse-Gamma (NIG) prior over the prediction targets. The network is then optimized using a comprehensive loss function that combines an evidential Negative Log-Likelihood and evidential regularization with deterministic L1 and MAPE penalties ( see Section \ref{sec:loss_evidential}). Further implementation details are available in the supplement, where we also justify our choice of GIN and compare it against alternative decoders.

\section{Training}
\label{sec:training}

\subsection{Datasets Preparation}
\begin{figure}[ht]
  \centering
  \includegraphics[width=0.77\textwidth]{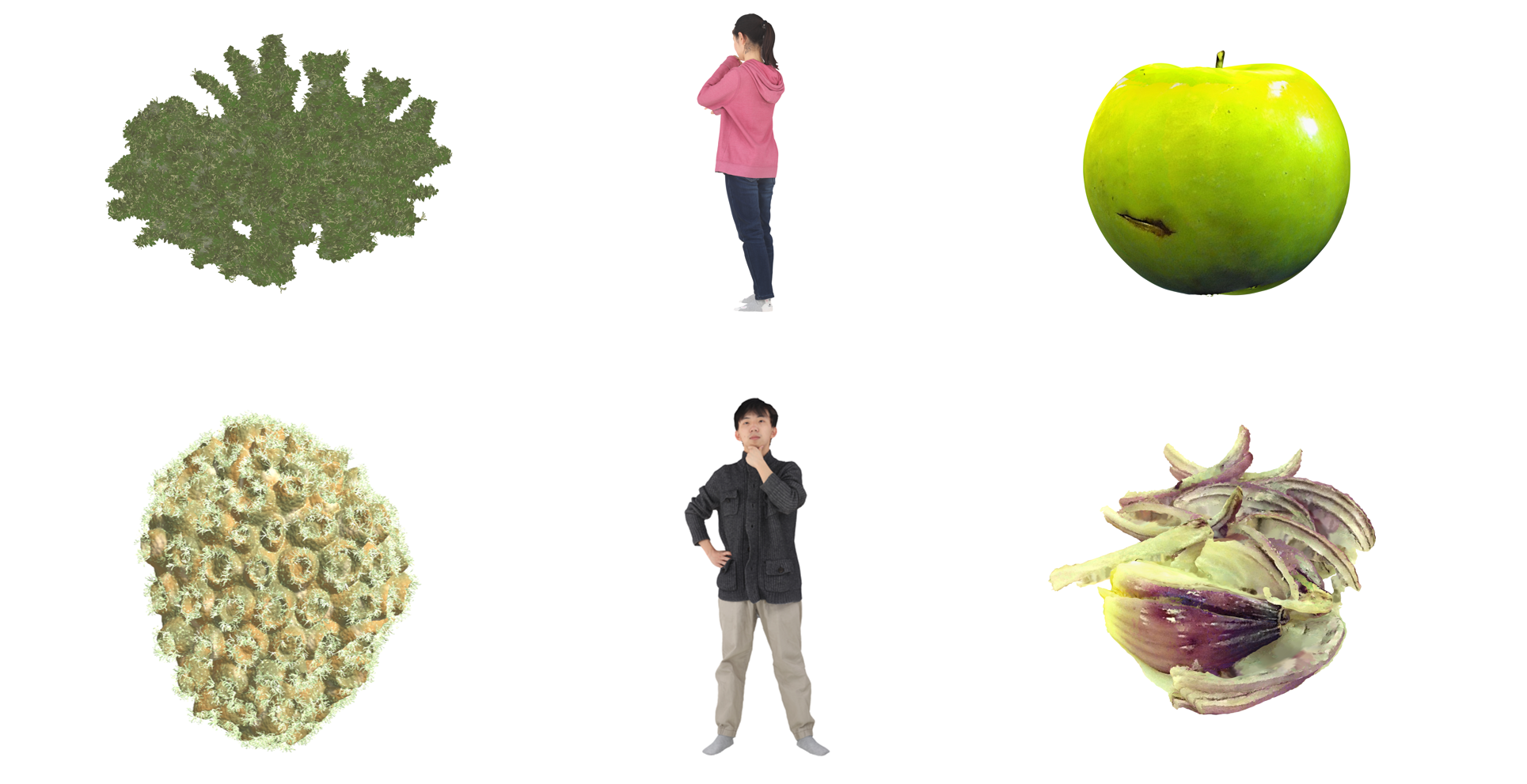}
  \caption{
\textbf{Representative samples across datasets}. 
Top: food items from the \textit{MetaFood} dataset. 
Middle: human subjects from \textit{THuman2.1}. 
Bottom: synthetic coral specimens generated with \textit{Infinigen}. 
These examples highlight the geometric and visual diversity of the domains on which our framework performs unified volume and surface estimation.}
  \label{fig:food_cor_hum}
\end{figure}

In our training pipeline, we first train a coarse version of the model on the Objaverse dataset for robust generalization, followed by fine-tuning on domain-specific datasets (\textbf{Synthetic Corals}, \textbf{THuman2.1}~\cite{tao2021function4d}, and \textbf{MetaFood}~\cite{chen2024metafood3d}) to adapt to particular applications. In Figure~\ref{fig:food_cor_hum}, we illustrate some assets from each of these datasets. This approach leverages Objaverse for initial broad learning and subsequent specialization for accurate volume and surface area estimation. Another motivation for this training approach is the lack of large datasets in the specific applications.

To ensure reliable ground-truth geometry, all 3D meshes across our datasets are first processed with ManifoldPlus~\cite{huang2020manifoldplus} to guarantee they are watertight.

\noindent \textbf{Objaverse:} For the Objaverse dataset, we select meshes and generate multi-view images through projections. We compute ground-truth (GT) volume and surface area metrics from these watertight meshes (as explained in the following), providing a strong foundation for the initial coarse training.

\noindent \textbf{Synthetic Corals:} Due to the lack of adequate 3D captured Coral dataset, we constructed our own dataset of textured coral meshes generated with Infinigen~\cite{raistrick2023infinite}. The dataset contains more than 4000 different assets, representing corals of different species, shape and size. We plan to release it upon acceptance. 

\noindent \textbf{THuman2.1:} Originally this dataset contains textured human meshes, that we use to evaluate the GT metrics. We apply a similar strategy of rendering masked multi-view images with black backgrounds on roughly 2500 assets and random view sampling as for the corals.  

\noindent \textbf{MetaFood:} This dataset is composed of real food photographs. We used segmented version of this images using Grounded SAM~2, prompted by the food class name. We perform manual quality assurance to retain accurate masks, correcting inconsistencies (e.g., varying inclusion of containers) to ensure alignment with ground-truth metrics before using them in training. The final dataset size consists of 360 assets.



\subsection{Loss Function}
\label{sec:loss_evidential}

To capture both predictive uncertainty and model confidence, we employ a \textit{Deep Evidential Regression} framework \cite{amini2020deep}. Unlike standard regression that outputs a point estimate or a fixed variance, we treat the target $y$ as being drawn from a Gaussian distribution $\mathcal{N}(\mu, \sigma^2)$ where the mean and variance are themselves unknown. We place a Normal-Inverse-Gamma (NIG) prior over these parameters:
\begin{equation}
( \mu, \sigma^2 ) \sim \text{NIG}(\gamma, v, \alpha, \beta)
\end{equation}

Our network is modified to output four parameters for each target: $\gamma$ (the prediction), $v$ (evidence scale), and $\alpha$, $\beta$ (scaling the shape and scale of the variance). The loss function $\mathcal{L}_{\text{tot}}$ consists of two primary components: a Negative Log-Likelihood ($\mathcal{L}_{\text{NLL}}$) term and an evidential regularizer ($\mathcal{L}_{\text{reg}}$).

\subsubsection{Evidential Negative Log-Likelihood}
By marginalizing out the unknown $\mu$ and $\sigma^2$, the model minimizes the NLL of the resulting Student-t distribution. Given the observed target $y$, the loss is defined as:

\begin{equation}\label{eq:NLL_evidential}
\begin{aligned}
\mathcal{L}_{\text{NLL}}(y; \gamma, v, \alpha, \beta) &= \frac{1}{2}\log\left(\frac{\pi}{v}\right) - \alpha\log\left( 2\beta(1 + v) \right) \\
&\hspace{-2.9em}+ \left(\alpha + \frac{1}{2}\right)\log\left((y - \gamma)^2 v  
+ 2\beta(1 + v) \right) 
+ \log\left(\frac{\Gamma(\alpha)}{\Gamma(\alpha + 1/2)}\right)
\end{aligned}
\end{equation}
where $\Gamma(\cdot)$ is the gamma function. This term rewards the model for aligning the predictive mean $\gamma$ with the target while adjusting the evidence parameters to explain the data dispersion.

\subsubsection{Evidential Regularization}
To penalize high "evidence" in regions where the model makes large prediction errors, a regularization term is included. This forces the model to decrease its confidence (lower $v$ and $\alpha$) when the prediction $\gamma$ deviates from the ground truth $y$:
\begin{equation}\label{eq:reg_evidential}
\mathcal{L}_{\text{reg}} = |y - \gamma| \cdot (2v + \alpha)
\end{equation}

\subsubsection{Deterministic Penalties (L1 and MAPE)}
To further constrain the predictive mean $\gamma$ and ensure robust point estimation, we incorporate Mean Absolute Error (L1) and Mean Absolute Percentage Error (MAPE). While the MAPE loss ensures proportional accuracy relative to the target's magnitude, which is particularly beneficial for targets with wide scale variations, relying uniquely on MAPE can be unstable and highly sensitive to outliers, especially when the target values are small. The L1 loss is therefore added to enforce absolute precision regardless of scale and to stabilize the training process:
\begin{equation}\label{eq:loss_l1}
\mathcal{L}_{\text{L1}} = |y - \gamma|
\end{equation}
\begin{equation}\label{eq:loss_mape}
\mathcal{L}_{\text{MAPE}} = \left| \frac{y - \gamma}{y} \right|
\end{equation}

\subsubsection{Total Objective}
The final objective is a weighted combination of the evidential terms and the deterministic regression penalties, controlled by their respective hyperparameters $\lambda_{\text{reg}}$, $\lambda_{\text{L1}}$, and $\lambda_{\text{MAPE}}$:
\begin{equation}\label{eq:loss_tot}
\mathcal{L}_{\text{tot}} = \mathcal{L}_{\text{NLL}} + \lambda_{\text{reg}} \cdot \mathcal{L}_{\text{reg}} + \lambda_{\text{L1}} \cdot \mathcal{L}_{\text{L1}} + \lambda_{\text{MAPE}} \cdot \mathcal{L}_{\text{MAPE}}
\end{equation}
In practice, this formulation allows the model to differentiate between \textbf{aleatoric uncertainty} ($\frac{\beta}{\alpha-1}$) and \textbf{epistemic uncertainty} ($\frac{\beta}{v(\alpha-1)}$), while the L1 and MAPE terms provide robust gradient signals to rapidly and accurately converge the predictive mean toward the ground truth.

\subsection{Ground Truth Surface and Volume Estimation}
\label{subsec:geom-metrics}
In the following, we describe the methods for computing ground-truth surface area and volume metrics across our datasets, ensuring consistency and accuracy for model training and evaluation. For all data we begin with preprocessing to handle mesh quality and scaling.

For watertight preprocessing, meshes from the \textbf{Objaverse}, \textbf{THuman2.1} and \textbf{Synthetic Corals} datasets are processed using ManifoldPlus~\cite{huang2020manifoldplus} to ensure watertightness, which is essential for reliable volume computations.


Normalization is applied uniformly across all datasets to ensure strict scale-invariance. To robustly handle potential floating artifacts or noise, we first perform statistical outlier removal on the mesh vertices, discarding points whose spatial positions deviate by more than three standard deviations from the mean. 
For the filtered vertices of a given triangular mesh $\mathcal{M}$, we compute its exact minimum bounding sphere. We then perform isotropic scaling such that the diameter of this minimal enclosing sphere is exactly 1. Letting $R$ denote the radius of the minimum bounding sphere, the scaling factor is defined as $s = 1 / (2R)$, resulting in the normalized mesh $\tilde{\mathcal{M}} = s\mathcal{M}$. All reported geometric metrics, including surface area and volume, are calculated based on this normalized form.

Surface area is calculated for all datasets summing the areas of individual triangles on the surface of the mesh, while we compute the signed volume for ground truth volume estimation. This methodology ensures accurate and dataset-appropriate ground-truth metrics, supporting the evaluation and refinement of our model. In the Supplements, we also mention other types of normalization and additional ground truth estimation methods for validation.

\section{Results}
\label{sec:results}

\subsection{Ablation Studies}
\label{sec:results-overview}
We conducted ablation studies to analyze the contribution of the 2D branch and the choice of multi-view 3D backbone (VGGT vs. MapAnything). 

Table~\ref{tab:mape_thuman_ablation} presents the Mean and Median Absolute Percentage Error (MAPE and MdAPE) for volume and surface on reconstruction performance on \textbf{THuman}, evaluating the effects of the inclusion or exclusion of the 2D branch and the backbone selection. 
Our results demonstrate that incorporating the 2D-encoded features significantly reduces errors for both volume and surface estimates. 

Table~\ref{tab:mape_vs_samples_thuman_corals_metafood} examines the model's robustness to training with varying numbers of samples and across three datasets (\textbf{MetaFood}, \textbf{THuman}, and \textbf{Synthetic Corals} ). For these experiments, MapAnything is used as the encoder due to its memory efficiency, allowing it to scale up to 200 input images while maintaining performance comparable to VGGT (see Table~\ref{tab:mape_thuman_ablation}).
Surface reconstruction consistently achieves lower MAPE and MdAPE values than volume reconstruction, indicating that it is inherently easier. Performance varies by dataset, with the model excelling on \textbf{THuman}, followed by \textbf{Synthetic Corals}, and then \textbf{MetaFood}. This variation can be attributed to several factors:
\begin{itemize}
\item The level of sample variability within each dataset;
\item Potential errors in ground-truth estimation derived from meshes (see the Supplements);
\item Differences in dataset quality;
\item Total number of samples available;
\item Small-volume objects amplifying percentage errors (further detail in Supplementary section).
\end{itemize}




\begin{table}[ht]
\centering
\scriptsize 
\setlength{\tabcolsep}{4pt} 
\renewcommand{\arraystretch}{0.7} 
\caption{Ablation study comparing the effects of DINOv3 features and backbone choices (VGGT vs. MapAnything) on MAPE and MdAPE for volume and surface reconstruction on the \textbf{THuman} dataset, using 30 images per asset.}
\label{tab:mape_thuman_ablation}
\begin{tabular}{ll cccc}
\toprule
\multicolumn{2}{c}{}
& \multicolumn{2}{c}{\textbf{Volume} }
& \multicolumn{2}{c}{\textbf{Surface} }\\
\cmidrule(lr){3-4}\cmidrule(lr){5-6}
\textbf{Model} & \textbf{Variant}
& MAPE $\downarrow$& MdAPE $\downarrow$& MAPE $\downarrow$& MdAPE $\downarrow$\\
\midrule  
VGGT        & w/ DINOv3  & 7.97  & 6.56  & 4.84  & 4.30 \\
VGGT        & w/o DINOv3 & 20.07 & 14.77 & 11.80 & 9.08 \\
MapAnything & w/ DINOv3  & 7.95  & 6.68  & 6.27  & 5.39 \\
MapAnything & w/o DINOv3 & 18.70 & 14.90 & 11.16 & 8.77 \\
\bottomrule
\end{tabular}
\end{table}

\begin{table}[t]
\centering
\scriptsize
\setlength{\tabcolsep}{3.0pt}      
\renewcommand{\arraystretch}{0.8}  
\caption{Robustness of Volume and Surface Estimation on three datasets: Mean and Median APE versus Number of Samples per Object (lower is better).}
\label{tab:mape_vs_samples_thuman_corals_metafood}
\begin{tabular}{ll cccccccc}
\toprule
\multirow{2}{*}{\textbf{Dataset}} & \multirow{2}{*}{\textbf{Metric}} & \multicolumn{8}{c}{\textbf{Number of Samples}} \\
\cmidrule(lr){3-10}
& & \textbf{1} & \textbf{2} & \textbf{4} & \textbf{8} & \textbf{16} & \textbf{32} & \textbf{64} & \textbf{100+} \\
\midrule
\multirow{4}{*}{\textbf{THuman}}
& Vol. MAPE $\downarrow$  & 10.97 & 9.96  & 8.60  & 6.98  & 6.93  & 6.63  & 6.30  & 6.25 \\
& Vol. MdAPE $\downarrow$ & 8.68  & 8.10  & 5.71  & 5.37  & 5.14  & 4.68  & 4.48  & 4.11 \\
& Surf. MAPE $\downarrow$ & 8.45  & 6.69  & 5.56  & 5.96  & 5.40  & 5.29  & 5.44  & 5.34 \\
& Surf. MdAPE $\downarrow$& 6.65  & 5.11  & 4.48  & 4.78  & 4.12  & 4.19  & 4.18  & 4.09 \\
\midrule
\multirow{4}{*}{\textbf{Corals}}
& Vol. MAPE $\downarrow$  & 19.83 & 16.83 & 15.28 & 13.68 & 12.44 & 11.86 & 12.07 & 11.88 \\
& Vol. MdAPE $\downarrow$ & 10.86 & 9.48  & 8.08  & 8.59  & 7.17  & 6.57  & 6.73  & 6.37 \\
& Surf. MAPE $\downarrow$ & 9.46  & 8.63  & 7.62  & 6.90  & 6.60  & 6.18  & 6.09  & 6.06 \\
& Surf. MdAPE $\downarrow$& 7.74  & 7.08  & 6.26  & 5.45  & 5.14  & 4.72  & 4.86  & 4.80 \\
\midrule
\multirow{4}{*}{\textbf{MetaFood}}
& Vol. MAPE $\downarrow$  & 47.03 & 38.78 & 49.92 & 35.90 & 34.20 & 38.42 & 32.38 & 29.68 \\
& Vol. MdAPE $\downarrow$ & 34.07 & 23.08 & 21.58 & 16.73 & 21.44 & 14.81 & 20.16 & 15.35 \\
& Surf. MAPE $\downarrow$ & 16.33 & 17.41 & 15.49 & 15.40 & 13.01 & 13.20 & 14.04 & 13.72 \\
& Surf. MdAPE $\downarrow$& 13.16 & 14.37 & 13.76 & 14.17 & 10.04 & 10.61 & 12.14 & 12.36 \\
\bottomrule
\end{tabular}
\end{table}

\subsection{Comparison to Food Volume Estimation Benchmarks}
\label{sec:food-volume-comparison}
We compare our method on the \textbf{MetaFood} dataset against established baselines for food volume estimation. We trained our model on subsets of \textbf{MetaFood} using $\NImg \in {1, 2, 4, 8, 16, 32, 64, 128, 200}$ images per sample, employing MapAnything for reconstruction. In addition, we also evaluated a decoder variant with $\lambda_{\text{L1}}$, and $\lambda_{\text{MAPE}}$ terms disabled.
We benchmarked our approach against baselines such as ININ and VolETA on the 20-scene protocol, evaluating performance across all samples as well as a challenging subset of five single-image cases (Table~\ref{tab:food_averages_wins}). Across the 18 directly comparable scenes, our model trained with both the evidential and deterministic losses achieved the lowest Mean Absolute Percentage Error (MAPE). Furthermore, our evidential-only variant secured the lowest overall Median APE (MdAPE), while also recording the lowest MAPE and MdAPE on the hardest single-view scenes. Thus, our approach excels both in multi-view and in the single-image setting, underscoring its robustness with sparse observations.
The model, in addition, successfully reconstructed two additional cases that prior studies omitted due to reported quality issues (see the per-scene breakdown in the supplementary material). We exclude VolTex results from this comparison.\footnote{VolTex reports results only on a subsample of the protocol, so a direct comparison is not appropriate.}

%
%
\subsection{Uncertainty Quantification}
\label{sec:uncertainty-quantification}

\begin{figure}[htbp]
  \centering
  \includegraphics[width=0.49\textwidth, height=6.7cm]{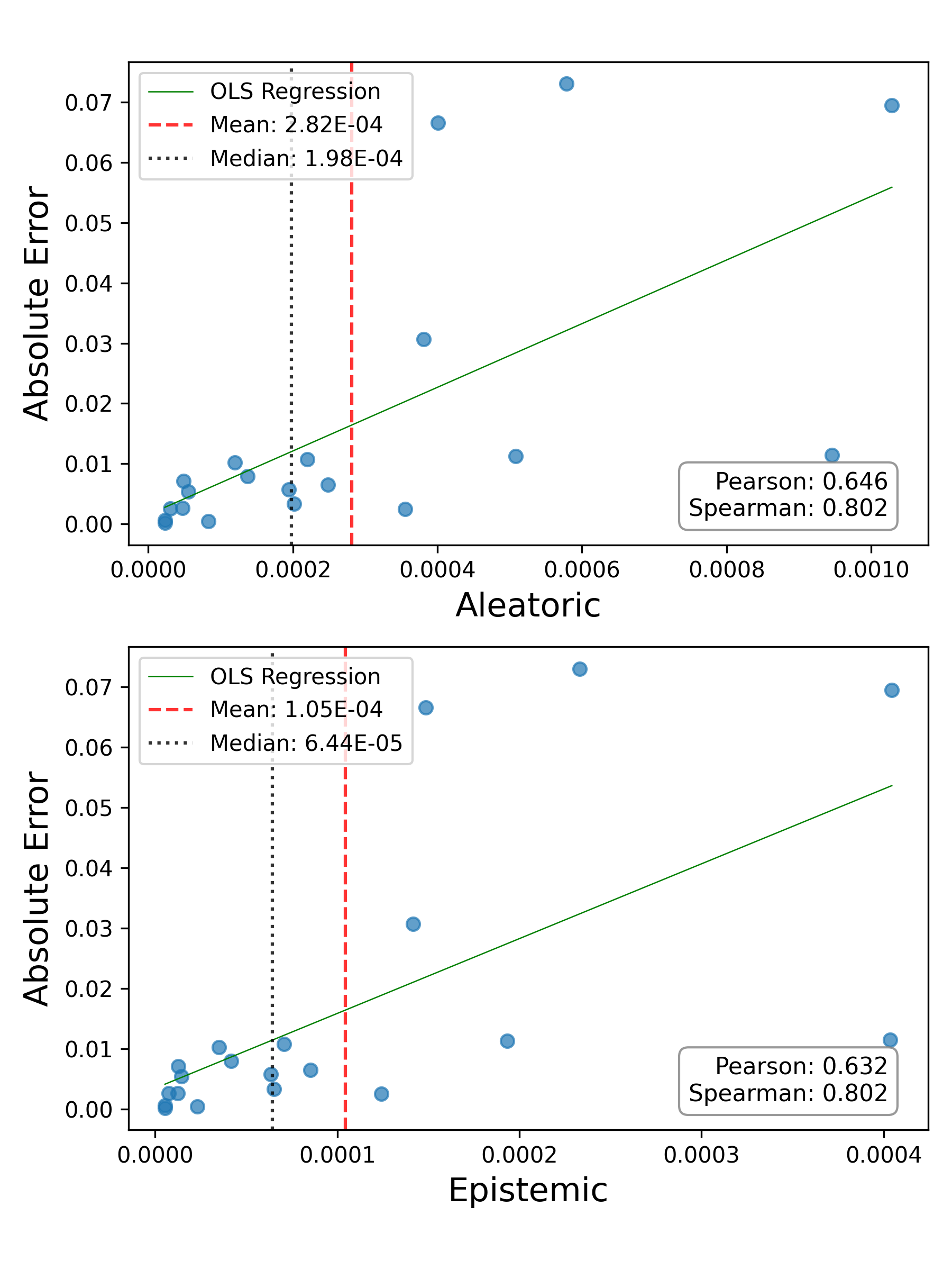}
  \hfill  
  \includegraphics[width=0.49\textwidth, height=6.7cm]{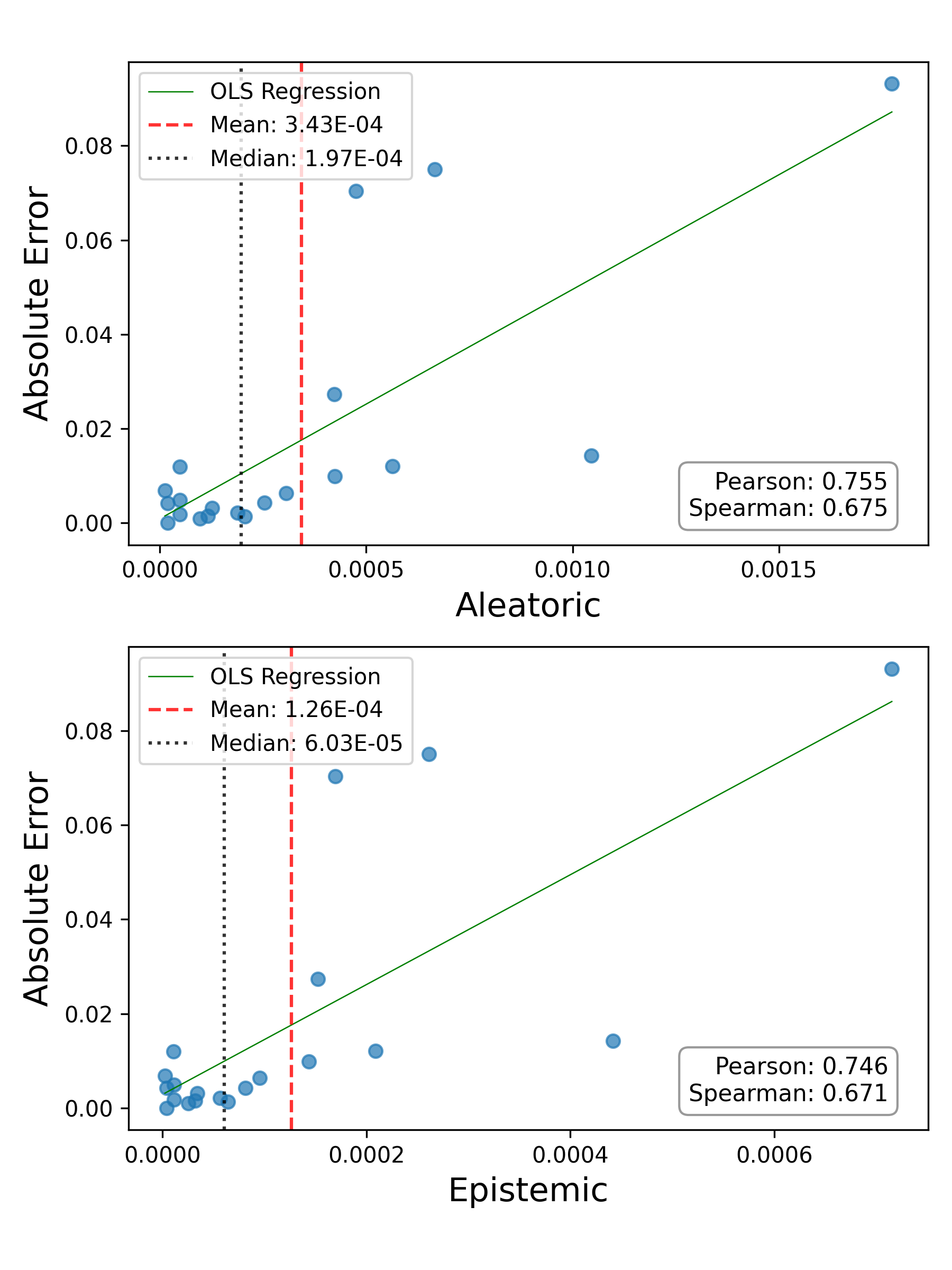}
  \caption{\textbf{Absolute Error vs.\ Uncertainty.} Correlation between absolute error and predicted aleatoric/epistemic uncertainty for Evidential (\emph{left}) and Evidential+Deterministic (\emph{right}) models. Higher uncertainty aligns with larger errors.}
  \label{fig:uncertainty_plots}
\end{figure}
\begin{figure}[htbp]
  \centering
  \includegraphics[width=0.85\textwidth]{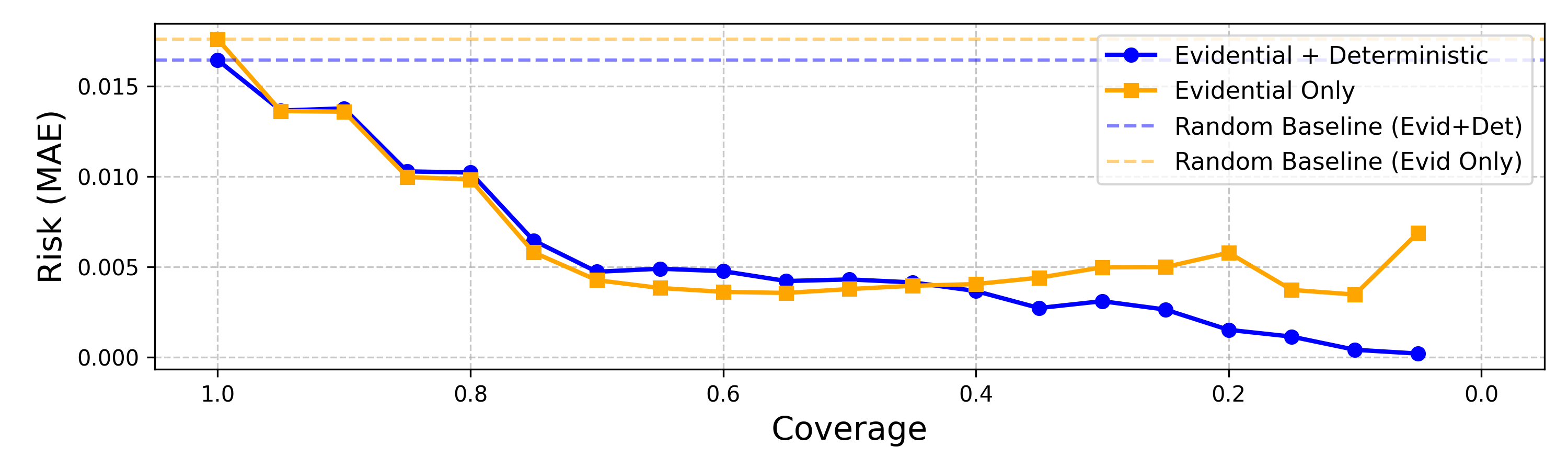}
  \caption{\textbf{Risk--Coverage (Total Uncertainty).} MAE decreases as high-uncertainty samples are rejected. Both models outperform random rejection; Evidential+Deterministic shows improved calibration at low coverage.}
  \label{fig:risk_coverage}
\end{figure}

To evaluate prediction reliability, we analyze the correlation between Absolute Error and predicted uncertainties on the 20-sample MetaFood test set (Figure~\ref{fig:uncertainty_plots}). Positive trends across both aleatoric and epistemic metrics indicate that higher predicted uncertainty effectively corresponds to higher actual reconstruction errors, and confirm that the model successfully flags unreliable predictions. Detailed per-sample uncertainties are provided in the supplementary material.
We further validate practical utility via Risk-Coverage analysis, tracking overall error (Risk) as highly uncertain predictions are incrementally rejected (Coverage). Figure~\ref{fig:risk_coverage} demonstrates that thresholding by combined uncertainty sharply reduces error compared to a random baseline. The evidential-only model drops risk by over 70\% up to 0.6 coverage, despite slight overconfidence at extremely low coverages. Incorporating the deterministic loss, instead, yields a curve that approaches near-zero error for the most confident predictions. This hybrid formulation improves overall calibration, enabling robust deployment where uncertain predictions are reliably flagged for human review.

\begin{table}[t]
\centering
\scriptsize
\caption{MetaFood Challenge: MAPE and MdAPE on the full dataset and single-image setting. We report our results with Evidential and Evidential+Deterministic losses.}
\label{tab:food_averages_wins}
\setlength{\tabcolsep}{3pt} 
\renewcommand{\arraystretch}{0.8} 
\begin{tabular}{l l ccccc}
\toprule
\multirow{2}{*}{\textbf{Scope}} & \multirow{2}{*}{\textbf{Metric}} & \textbf{Ours} & \textbf{Ours} & \multirow{2}{*}{\textbf{ININ}} & \multirow{2}{*}{\textbf{VolETA}} & \multirow{2}{*}{\textbf{Trellis}} \\
 & & \textbf{(Evid.)} & \textbf{(Evid. + Determ)} & & & \\
\midrule
\multirow{3}{*}{All cases} 
& MAPE $\downarrow$   & 11.65 & \textbf{10.27}& 11.73 & 10.97 &  -\\
& MdAPE $\downarrow$ & \textbf{6.60}  & 7.06 & 9.04  & 6.93  &  -\\
\midrule
\multirow{3}{*}{Single-image} 
& MAPE $\downarrow$   & \textbf{14.74} & 15.18 & 19.16 & 19.12 & 238.20 \\
& MdAPE $\downarrow$ & \textbf{5.97}  & 12.29 & 20.03 & 23.42 & 92.40 \\
\bottomrule
\end{tabular}
\end{table}

\begin{table}[t]
    \centering
    \scriptsize
    \setlength{\tabcolsep}{3pt}
    \renewcommand{\arraystretch}{1.01}
    
    \begin{minipage}[t]{0.56\textwidth}
        \centering
        \caption{Real corals and THuman vs. 2DGS/Agisoft (20/30 views).}
        \label{tab:corals_thuman_2dgs_agisoft}
        \resizebox{\linewidth}{!}{%
        \begin{tabular}{l l l cccc}
        \toprule
        Dataset & Views & Stat. & \multicolumn{2}{c}{\textbf{Ours}} & \multicolumn{2}{c}{\textbf{2DGS/Agisoft}} \\
        \cmidrule(lr){4-5} \cmidrule(lr){6-7}
        & & & Vol & Surf & Vol & Surf \\
        \midrule
        \multirow{4}{*}{Corals}
        & \multirow{2}{*}{20} 
        & MAPE $\downarrow$  & \textbf{14.23} & \textbf{14.89} & 21.30 & 101.89 \\
        & & MdAPE $\downarrow$ & \textbf{11.33} & \textbf{5.96}  & 16.75 & 27.97  \\
        & \multirow{2}{*}{30} 
        & MAPE $\downarrow$  & \textbf{16.46} & \textbf{14.11} & 21.05 & 93.08  \\
        & & MdAPE $\downarrow$ & \textbf{10.70} & \textbf{5.83}  & 15.54 & 31.76  \\
        \midrule
        \multirow{4}{*}{THuman}
        & \multirow{2}{*}{20}
        & MAPE $\downarrow$  & \textbf{9.65}  & \textbf{12.00} & 18.64 & 46.98 \\
        & & MdAPE $\downarrow$ & \textbf{7.57}  & \textbf{11.27} & 18.04 & 49.16 \\
        & \multirow{2}{*}{30}
        & MAPE $\downarrow$  & \textbf{11.79} & \textbf{12.62} & 13.45 & 38.12 \\
        & & MdAPE $\downarrow$ & \textbf{10.39} & \textbf{12.18} & 14.42 & 35.85 \\
        \bottomrule
        \end{tabular}%
        }
    \end{minipage}\hfill
    \begin{minipage}[t]{0.42\textwidth}
        \centering
        \caption{Real corals and THuman vs. Trellis (single-view).}
        \label{tab:corals_thuman_trellis}
        \resizebox{\linewidth}{!}{%
        \begin{tabular}{l l cccc}
        \toprule
        Dataset & Stat. & \multicolumn{2}{c}{\textbf{Ours}} & \multicolumn{2}{c}{\textbf{Trellis}} \\
        \cmidrule(lr){3-4} \cmidrule(lr){5-6}
        & & Vol & Surf & Vol & Surf \\
        \midrule
        \multirow{2}{*}{Corals}
        & MAPE $\downarrow$  & \textbf{20.68} & \textbf{30.07} & 66.09 & 77.24 \\
        & MdAPE $\downarrow$ & \textbf{9.37}  & \textbf{13.73} & 90.43 & 71.24 \\
        \midrule
        \multirow{2}{*}{THuman}
        & MAPE $\downarrow$  & \textbf{20.72} & \textbf{15.17} & 72.24 & 43.91 \\
        & MdAPE $\downarrow$ & \textbf{15.42} & \textbf{12.14} & 73.55 & 47.33 \\
        \bottomrule
        \end{tabular}%
        }
    \end{minipage}
\end{table}

\subsection{Generalization Evaluation: Comparisons with 2DGS on Real Corals and Agisoft on THuman Dataset}
\label{sec:corals_real_2dgs_Agisoft}

To evaluate our model's real-world generalization\footnote{For the experiments in \ref{sec:corals_real_2dgs_Agisoft}--\ref{sec:corals_real_2dgs} (Tables~\ref{tab:corals_thuman_2dgs_agisoft}--\ref{tab:corals_thuman_trellis}), we use a DGCNN point-cloud encoder instead of GIN.}, we tested it on a small set of real coral samples and the THuman dataset, benchmarking against 2D Gaussian Splatting~\cite{huang20242d} (2DGS), and a commercial 3D reconstruction software: Metashape Agisoft. For the corals, we collected and photographed a limited number of physical specimens, generating multi-view image sets with varying capture counts. Due to 2DGS's reconstruction constraints, reliable results were only achievable with 20 or 30 input views, which are highlighted in Table~\ref{tab:corals_thuman_2dgs_agisoft}.

We fine-tuned our model on this real-coral subset, starting from weights pre-trained on synthetic corals using MapAnything and DINOv3 features. Despite the limited dataset of only 11 real samples, the fine-tuned model consistently outperformed 2DGS in both mean and median errors across all tested image counts. Additionally, as shown in Table \ref{tab:corals_thuman_2dgs_agisoft}  our framework surpassed Agisoft's performance on a subsample of the THuman dataset. These results underscore the proposed method's strong generalization capabilities under data scarcity, with potential for further enhancements as more annotated real samples are acquired. Examples of the collected real corals are available in the supplementary material.

\subsection{Single-View Performance: Comparison with Trellis on Real Corals and THuman}
\label{sec:corals_real_2dgs}

We extended our evaluation to assess our framework's performance against Trellis on the same real-coral and THuman samples in a single-view context. For each object, we processed 30 individual images and derived a volume estimate from each. The same procedure was applied to Trellis, where volume estimates were obtained directly from its reconstructed meshes using a single image generation. As detailed in Table~\ref{tab:corals_thuman_trellis}, our framework surpasses Trellis across all metrics on both datasets.

\subsection{Runtime and Practicality}
Our model is \textbf{lightweight} and designed for deployment: considering a 50k-point mesh, it performs the forward pass in the fusion decoder in less than 0.1 seconds on average on an NVIDIA A100 GPU, with the GIN decoder requiring only 3024 Megabytes of memory. In contrast, reconstruction-based pipelines typically incur substantially higher runtimes, needing several minutes (and even up to an hour) to complete the reconstruction due to per-scene optimization or heavy rendering. Beyond speed and memory efficiency, our architecture eliminates the need for watertight meshes or hand-crafted geometric formulas. This enables strong generalizability across datasets and flexibility with input images—from single views to many—while delivering comparable or superior accuracy on the benchmark.
\section{Conclusions}
\label{sec:conclusions}
We present a lightweight multi-view framework that integrates 3D point clouds with pooled DINOv3 features through a compact graph decoder, facilitating direct and efficient regression of volume and surface area. Our model achieves low MAPE and MdAPE across diverse domains, including foods, humans, and corals, using a single forward pass, which provides significant speed benefits over 3D reconstruction-based methods and demonstrates robustness with sparse views; even in single-image scenarios, it outperforms existing baselines. Despite these performances, limitations arise from the need for scale-normalized predictions, as training on unit-extent geometry requires an external scaling cue, such as reference objects like utensils or markers, to convert estimates to absolute measurements, though the limited availability of datasets incorporating such references poses a substantial challenge for broader adoption. Furthermore, while the model's data efficiency is evident from its strong performance on the MetaFood dataset with only 360 samples or on real corals with few samples, opportunities for enhancement exist through the expansion of curated datasets, which could improve accuracy and strengthen cross-domain generalization by addressing gaps in representation and increasing the diversity of training examples. These limitations underscore valuable avenues for future research, positioning the framework to realize even greater precision and adaptability in practical applications.

\bibliographystyle{splncs04}
\bibliography{main}

\setcounter{page}{1}
\appendix

%
%
%

\section{Datasets Preparation Details}
In this section, we describe the data generation and preprocessing pipelines used to create our training and evaluation datasets: Objaverse~\cite{deitke2023objaverse} (for pre-training), synthetic corals, human bodies (THuman 2.1~\cite{tao2021function4d}), and real food items (CVPR MetaFood~\cite{chen2024metafood3d}).


\subsection{Objaverse}
\begin{figure}[htbp]
    \centering
    \includegraphics[width=\linewidth]{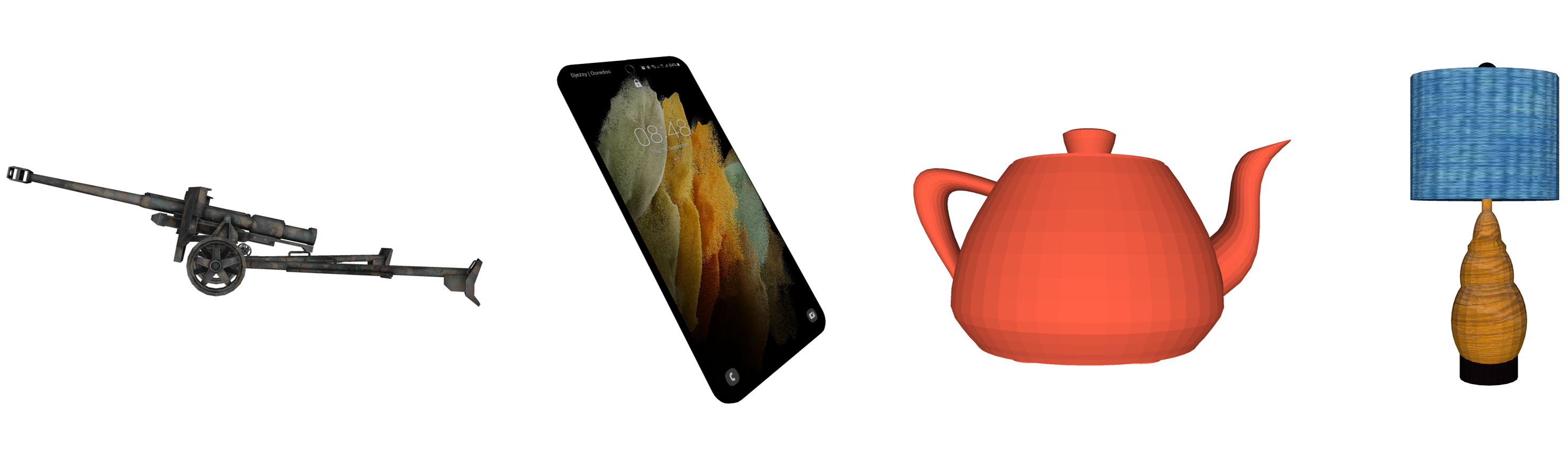}
    \caption{Examples of 3D assets from the Objaverse dataset used in our pipeline, illustrating the diversity of shapes and textures.}
    \label{fig:objaverse_objs}
\end{figure}

For Objaverse, we start from the textured Objaverse meshes converted to OBJ format, retaining both a raw textured version for rendering and a watertight counterpart for reliable geometry measurements. The dataset consists of $\approx 32{,}000$ assets (see Figure~\ref{fig:objaverse_objs}). For each object, we load the mesh and translate it to the origin. Ground-truth volume and surface area are computed from the watertight mesh using Trimesh, as detailed in the main paper. Multi-view RGB images are rendered off-screen in PyVista on a black background at $\approx 682 \times 512$ resolution, sampling $40$ camera poses per object with azimuth uniformly in $[0^\circ, 360^\circ)$, elevation restricted away from the poles ($|{\rm elev}| < 84^\circ$), distance in $(3.0\text{--}3.7)\times$ the object extent (with a mild bias toward closer views), and focal-point jitter $\pm 0.03\times$ the extent. The field of view is randomized in $24^\circ\text{--}28^\circ$ and the up-vector is fixed to $[0,0,1]$. Lighting uses $25\text{--}35$ scene lights with evenly spaced azimuth, random elevations in $[-80^\circ, 80^\circ]$, and intensity $0.13\text{--}0.16$, while PBR shading parameters are lightly randomized (ambient $0.15\text{--}0.30$, diffuse $0.5\text{--}0.7$, specular $0.15\text{--}0.3$, specular power $5\text{--}10$). For each rendered view, we save the RGB image and the corresponding camera parameters (position, quaternion, distance, angles, jitter, and FoV).

\subsection{Synthetic and Real Corals}
\begin{figure*}[t]
  \centering

  \begin{subfigure}{0.32\textwidth}
    \centering
    \includegraphics[width=\linewidth]{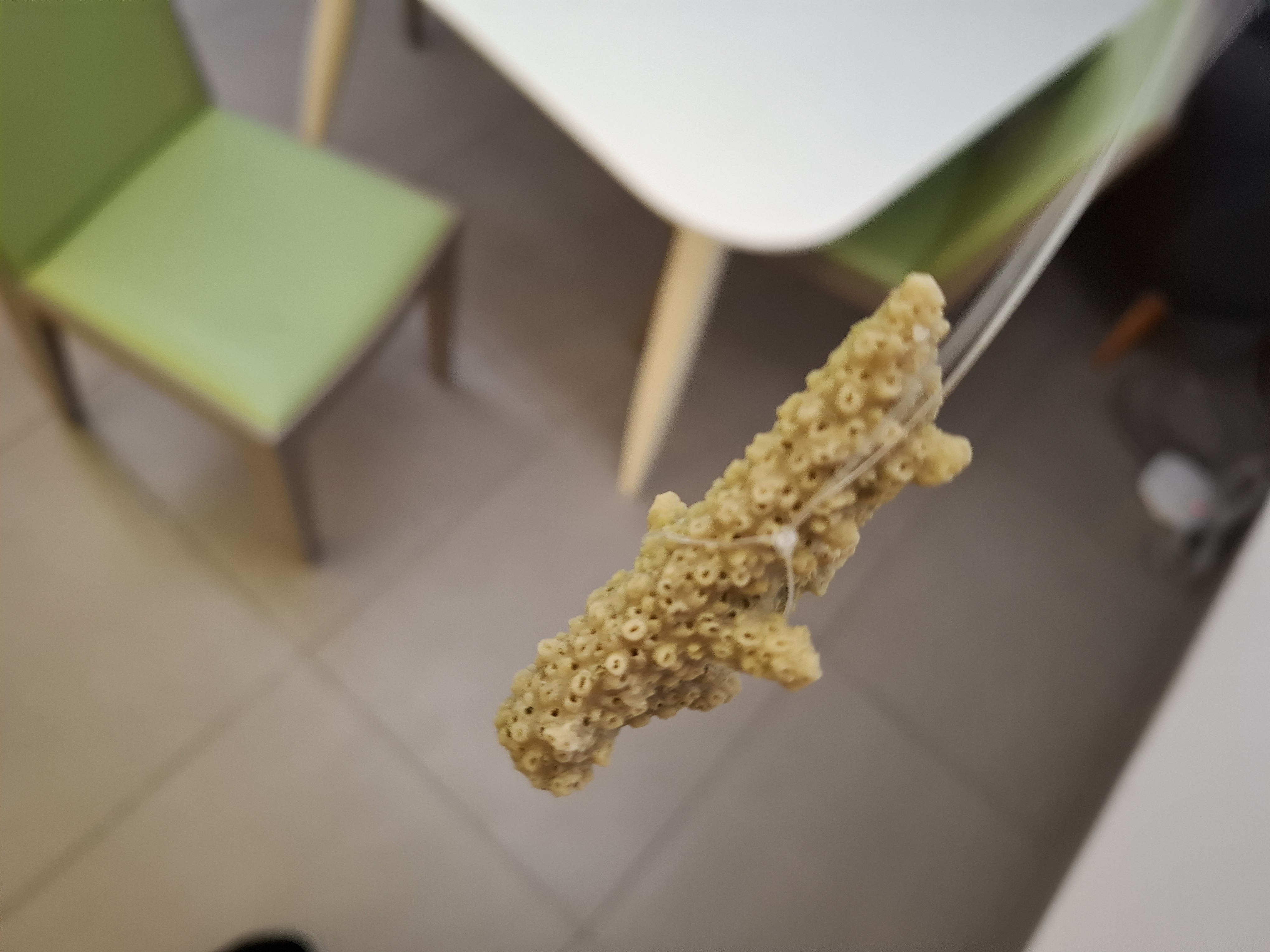}
    \caption{}
  \end{subfigure}
  \hfill
  \begin{subfigure}{0.32\textwidth}
    \centering
    \includegraphics[width=\linewidth]{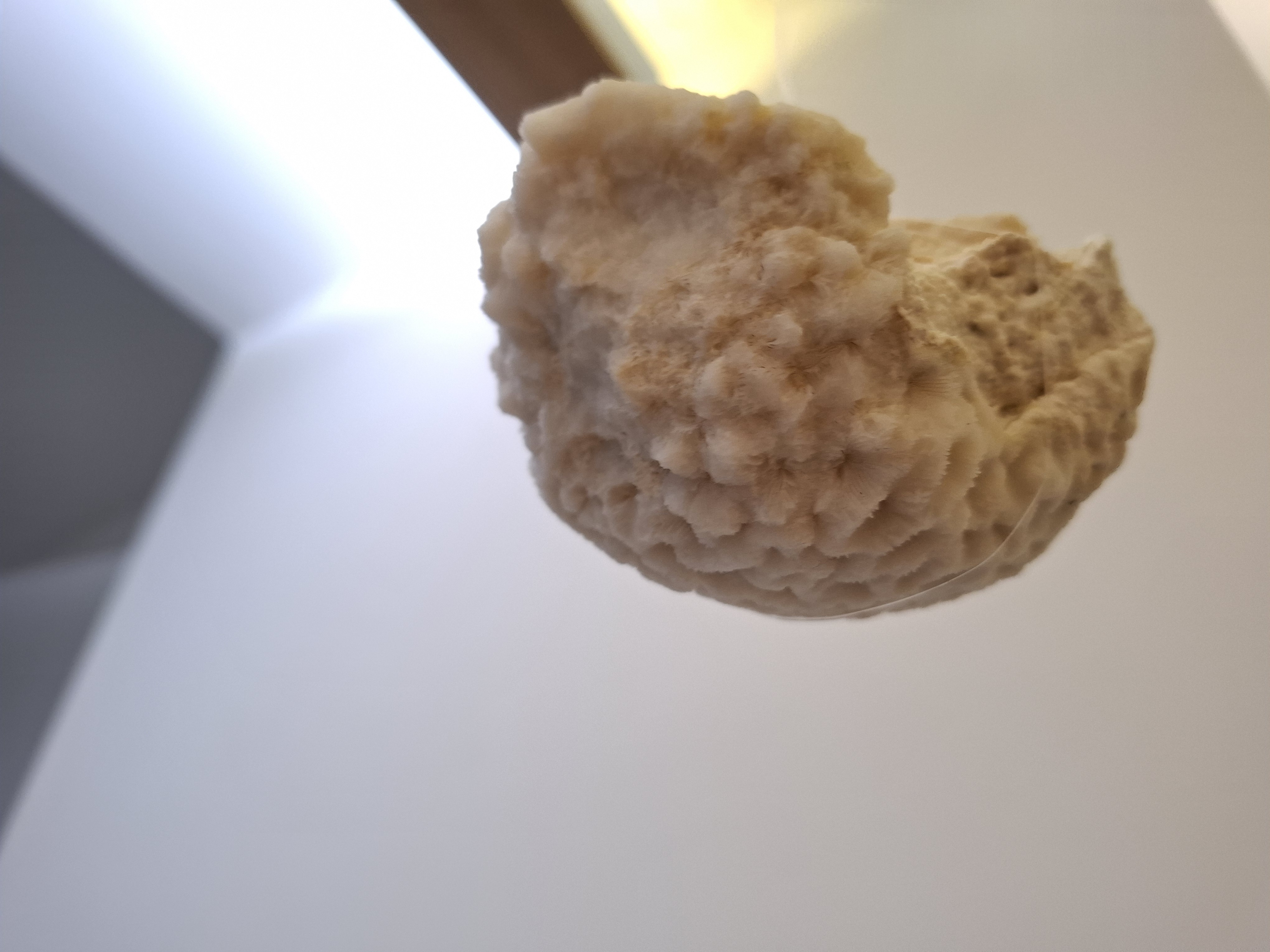}
    \caption{}
  \end{subfigure}
  \hfill
  \begin{subfigure}{0.32\textwidth}
    \centering
    \includegraphics[width=\linewidth]{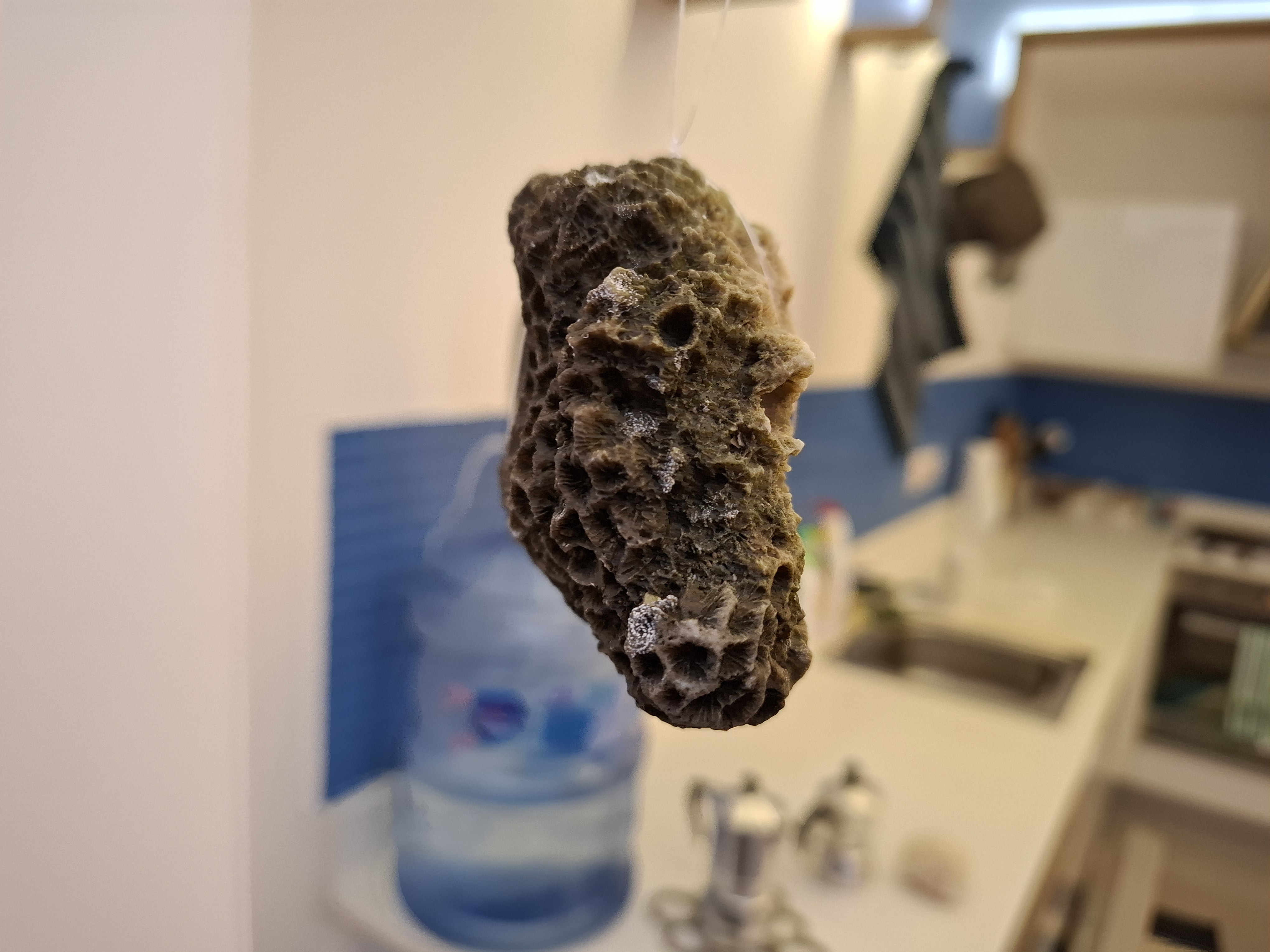}
    \caption{}
  \end{subfigure}

  \vspace{0.6em}

  \begin{subfigure}{0.32\textwidth}
    \centering
    \includegraphics[width=\linewidth]{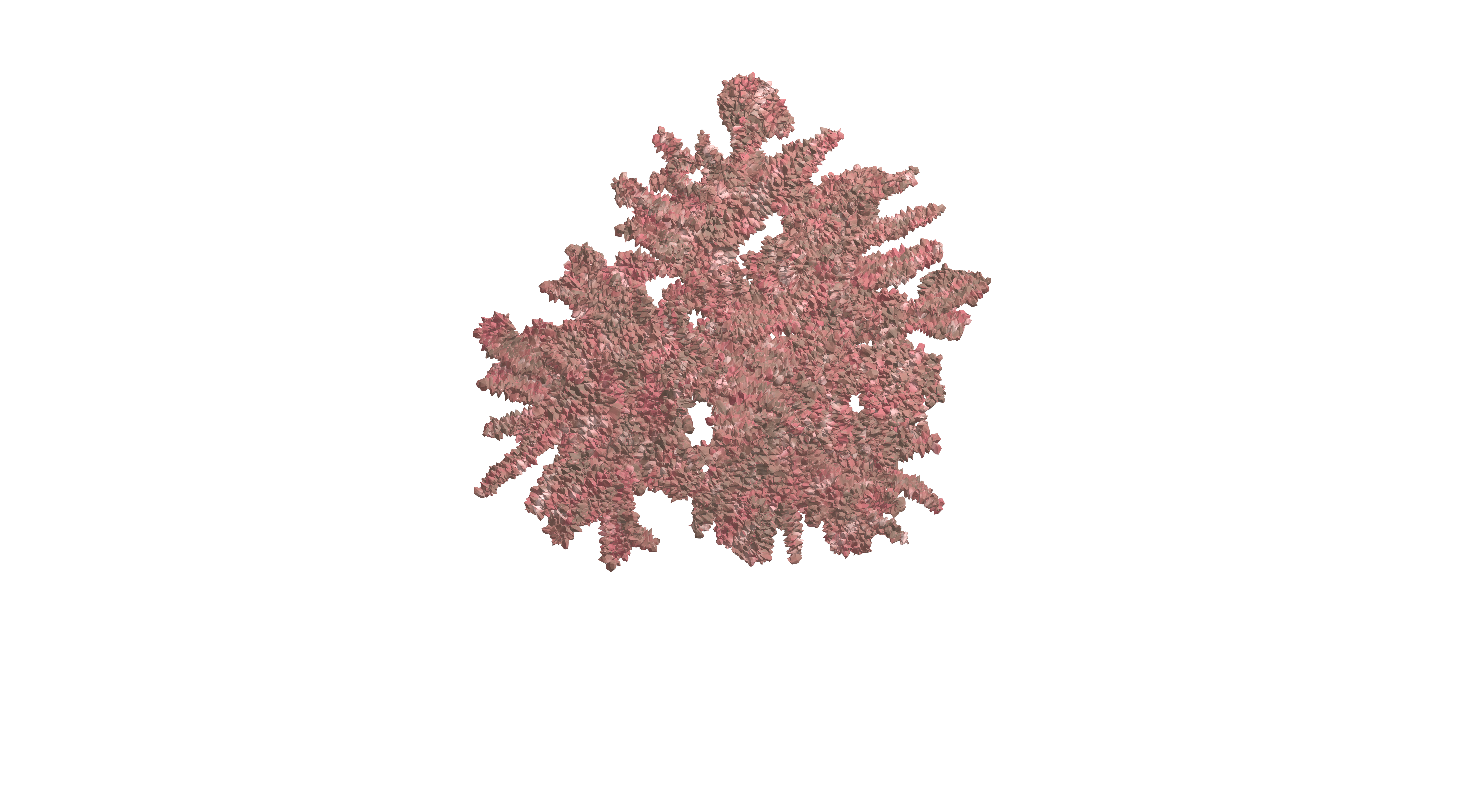}
    \caption{}
  \end{subfigure}
  \hfill
  \begin{subfigure}{0.32\textwidth}
    \centering
    \includegraphics[width=\linewidth]{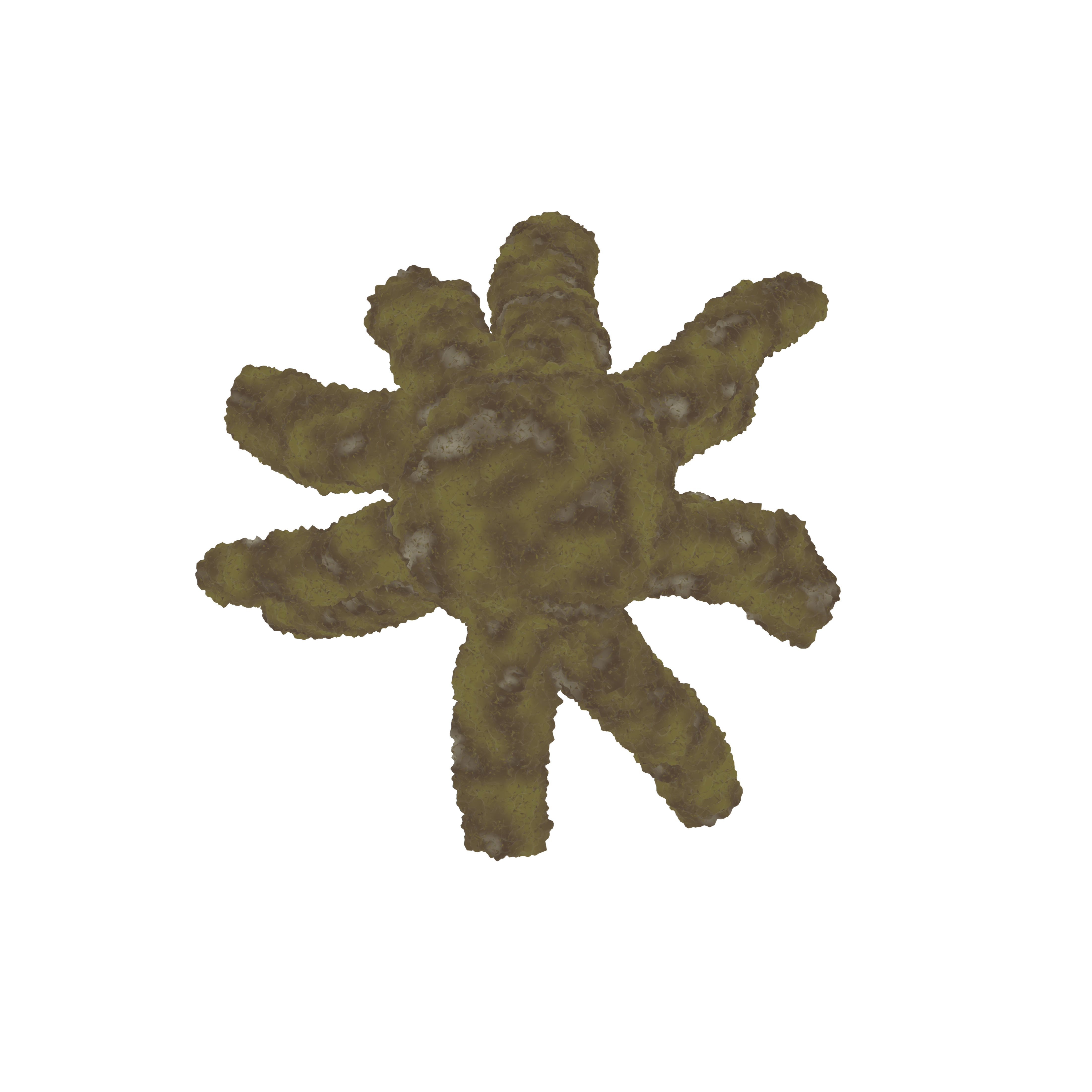}
    \caption{}
  \end{subfigure}
  \hfill
  \begin{subfigure}{0.32\textwidth}
    \centering
    \includegraphics[width=\linewidth]{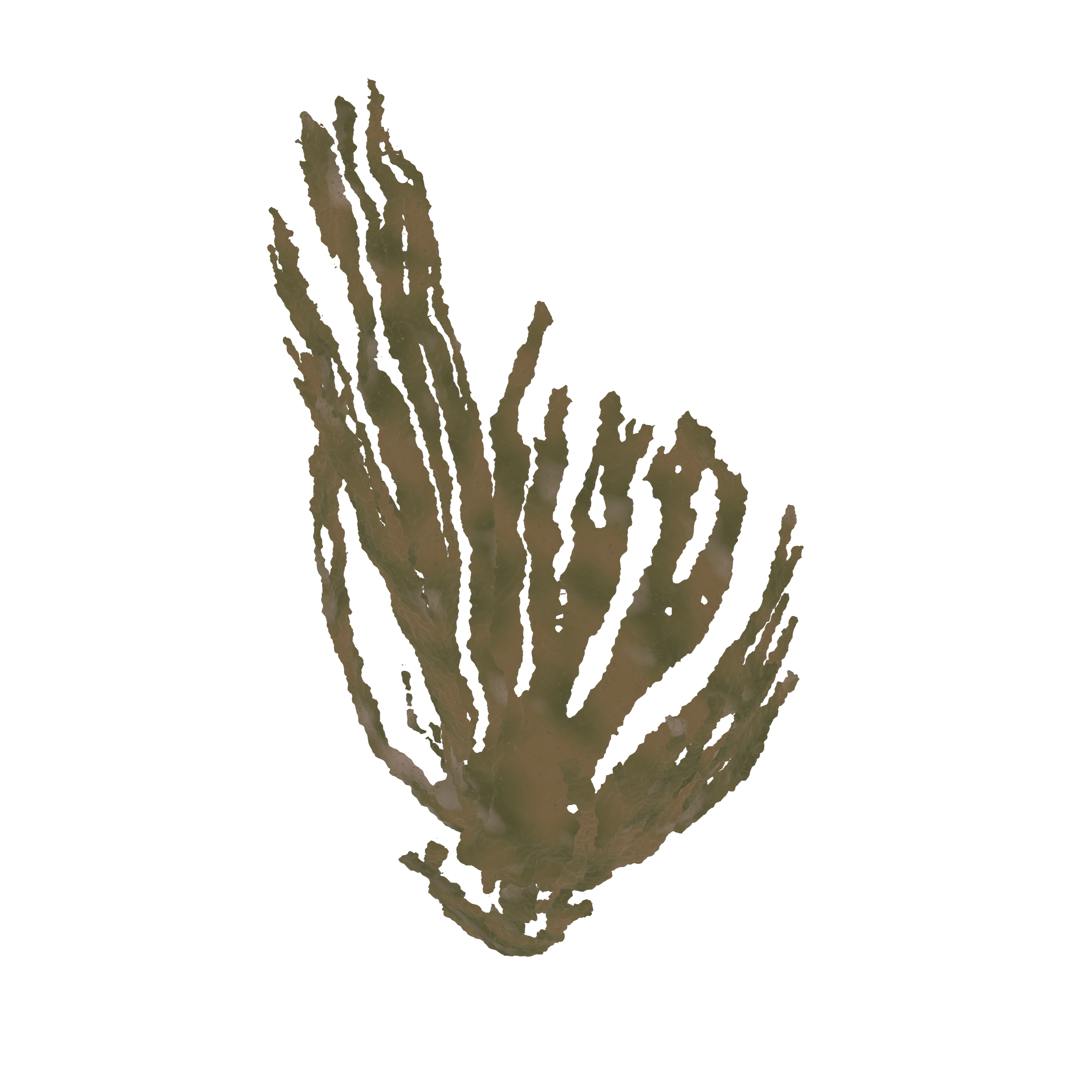}
    \caption{}
  \end{subfigure}

  \caption{Images of collected real corals (first row), and snapshots of synthetic corals (second row).}
  \label{fig:coral_examples_two_rows}
\end{figure*}

As mentioned in the main paper, using Infinigen~\cite{raistrick2023infinite}, we generated a synthetic coral dataset, that contains textured meshes of more than $4000$ different corals, with substantial variability in terms of represented species and size. From this dataset, we render multi-view RGB images using PyVista. Each rendered view has a resolution of \(1024\times768\) pixels with a field of view ranging from $22^{\circ}$ to $40^{\circ}$. We apply a slight random camera jitter to avoid pointing exactly at the object center; backgrounds are removed at render time, producing black-padded, centered images with a small random translation. Lighting uses \(32\) scene lights evenly spaced over \(360^\circ\) azimuth, elevations \(10^\circ\text{--}80^\circ\), intensity \(0.18\); shading is PBR-style with ambient \(0.2\), diffuse \(0.9\), specular \(0.4\), and full opacity. This setup ensures varied lighting and viewpoints while maintaining object focus.
Note that this process is performed before applying \textbf{ManifoldPlus}~\cite{huang2020manifoldplus} to produce watertight (textureless) geometry. The watertight geometry is solely used for ground-truth volume and surface area computation.\\
For real corals, we physically suspended each specimen on a thin cable and captured images all around it, collecting over 100 RGB views per coral (see Fig.~\ref{fig:coral_examples_two_rows} for real and synthetic corals utilized in our experiments). During fine-tuning, we vary the number of input images from 1 to 40 by randomly sampling subsets from these 100+ views for each coral.

\subsection{THuman~2.1}
The Thuman 2.1~\cite{tao2021function4d}  is a dataset of high-quality human scans captured using a dense DSLR rig. Each scan contains a 3D model with its corresponding texture map. Similarly to the synthetic corals dataset, we adopted a strategy of rendering images with black backgrounds and random view sampling. We render images in PyVista following the Objaverse protocol. We also adopt the same lighting parameters. The PBR shading parameters are as follows: (ambient \(0.22\text{--}0.33\), diffuse \(0.5\text{--}0.7\), specular \(0.16\text{--}0.3\), specular power \(5\text{--}10\), and full opacity).

\subsection{Food (CVPR MetaFood)}
For the CVPR MetaFood dataset~\cite{chen2024metafood3d}, we focused on real-world images and applied segmentation to isolate food items, incorporating manual quality assessment to address potential inconsistencies.

\begin{figure}[ht]
  
  \centering
  
  \includegraphics[width=0.41\textwidth]{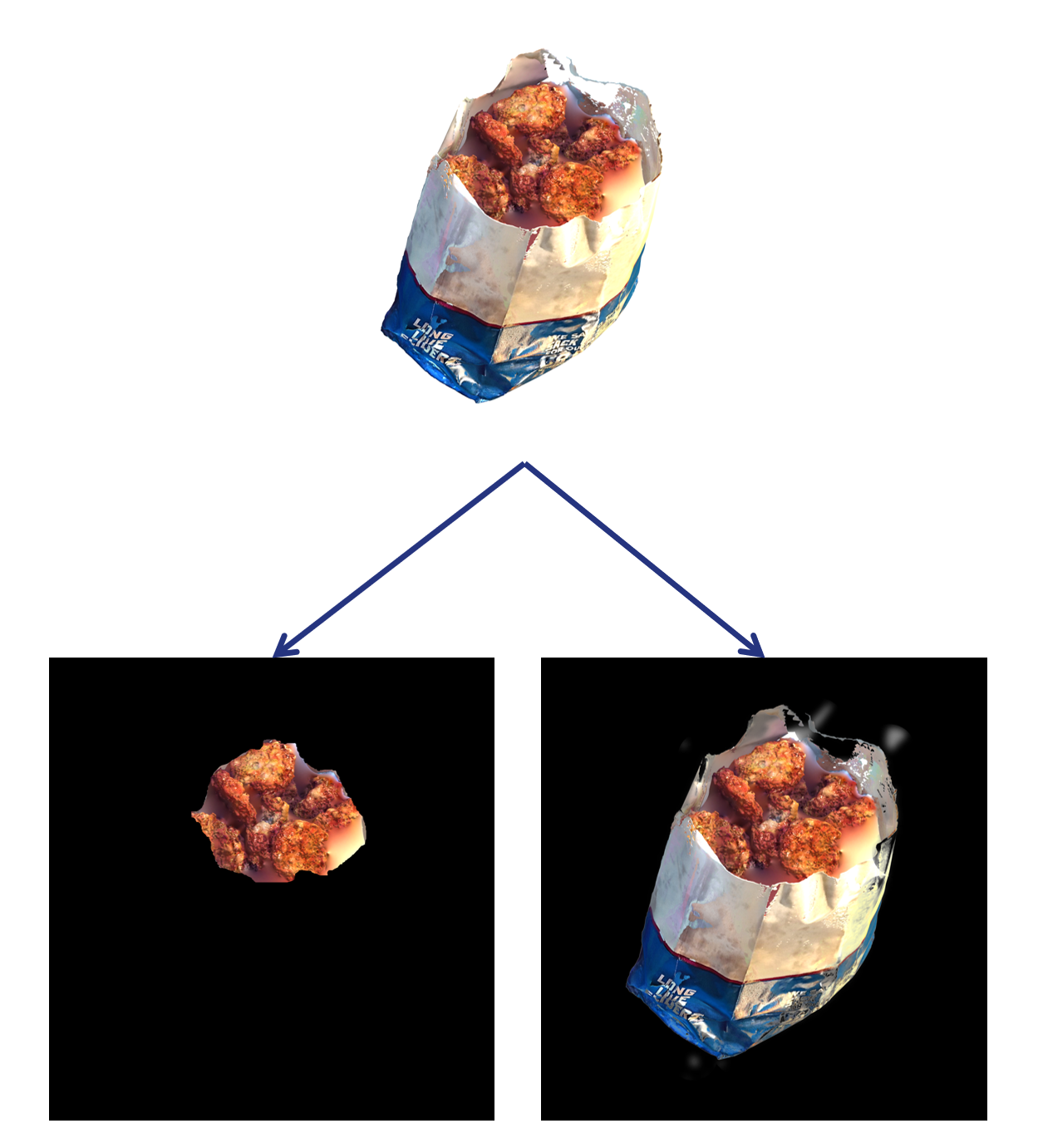}
\caption{
Example of segmentation inconsistencies in the MetaFood dataset. 
While some masks correctly isolate the full object (right), others segment only the food portion inside a container (left). 
This ambiguity arises because, in certain classes, the ground-truth volume refers to both the container and its contents, while the mask includes only the food. 
Manual quality assurance (QA) was therefore required to ensure consistency across samples. 
Representative segmentation examples were generated using Grounded-SAM2.
}
 \label{fig:masks_different3}
\end{figure}

Specifically, we used Grounded-SAM2 (combining Grounding DINO~\cite{simeoni2025dinov3} and SAM~2 for open-world, promptable segmentation) ~\cite{ren2024grounded,liu2023grounding,kirillov2023segany,ravi2024sam2segmentimages,ren2024grounding,jiang2024trex2}, by prompting the model with the food class name. Only masks that accurately segmented the food items were retained after manual review, ensuring the final inputs aligned with ground-truth assets. As discussed in the Methods section in the main paper and depicted in Figure~\ref{fig:masks_different3}, some classes in the MetaFood dataset exhibit ambiguities between the annotated volume (container + food) and the segmented region (food only). Such cases motivated the manual quality assessment step, where we filtered the masks to ensure that the supervision is consistent with the ground-truth assets.

\section{Network Architecture and Training Details}
\label{sec:supp-arch-training}

\subsection{Overall design.}
Our regressor operates on precomputed 3D point clouds with per-point features and DINO descriptors. For each object, we use a point-wise feature tensor of dimension 4 (3D coordinates plus one confidence scalar attribute) and a per-point DINO feature of dimension \(d_{\text{DINO}}=384\). The network has two main branches: a 3D branch (MapAnything~\cite{keetha2025mapanything}) and a 2D branch (DINO-feature encoder), followed by a Graph Isomorphism Network as the output of the 3D branch, fusion MLP and a small regression head that predicts the target scalar (volume or surface area) and its corresponding confidence. We use MapAnything as our default 3D reconstructor since it is computationally faster in our setup and remains stable in GPU memory when scaling to many input views, whereas VGGT often triggers out-of-memory errors in these regimes.

\subsection{3D map encoder.}
The 3D point cloud encoder is built using a GIN-style architecture to ensure high expressive power in capturing geometric structures. We define the graph using a $k$-nearest neighbor ($k$-NN) approach with $k=30$. The branch consists of five successive GIN layers with output dimensions of 64, 64, 128, 256, and 256, respectively. Each GIN layer follows the formal update rule:
\begin{equation}
    h_i^{(\ell+1)} = \text{MLP}^{(\ell)} \left( (1 + \epsilon^{(\ell)}) \cdot h_i^{(\ell)} + \sum_{j \in \mathcal{N}(i)} h_j^{(\ell)} \right)
\end{equation}
where 
\begin{itemize}
    \item \textbf{$h_i^{(\ell)}$ (Latent Feature):} Represents the hidden state of point $i$ at layer $\ell$, encoding its current local shape representation.
    \item \textbf{$\sum_{j \in \mathcal{N}(i)} h_j^{(\ell)}$ (Sum Aggregation):} Aggregates features from the $k$-nearest neighbors $\mathcal{N}(i)$. This term allows the model to better distinguish between varying neighborhood topologies and point densities.
    \item \textbf{$\epsilon^{(\ell)}$ (Learnable Scalar):} A layer-specific parameter that weights the importance of the central point's own features relative to its neighbors, preventing its unique geometric identity from being over-smoothed by the local context.
    \item \textbf{$\text{MLP}^{(\ell)}$ (Non-linear Mapping):} A multi-layer perceptron that projects the aggregated signal into a higher-dimensional space.
\end{itemize} 

Following the original GIN framework, we concatenate the output features from all five layers to form a multi-scale representation of 768 dimensions per point. A global adaptive max-pooling operation is then applied across the point dimension to extract a global shape descriptor. This descriptor is further refined through three fully connected layers ($768 \rightarrow 512$, $512 \rightarrow 256$, and $256 \rightarrow 128$), each utilizing \texttt{LayerNorm}, \texttt{LeakyReLU} (negative slope 0.2), and a dropout rate of 0.3, resulting in a 128-dimensional geometric embedding $z_{3D}$.



\subsection{2D feature branch.}
In parallel, we process the per-image DINO features through a two-layer MLP with hidden size 256 and output size 128. Each layer is followed by \texttt{LayerNorm}, LeakyReLU ($0.2$), and dropout ($0.3$). This yields a 128-dimensional feature for each image. We aggregate these features across different views using both max-pooling and mean-pooling and concatenate the two pooled vectors, resulting in a 256-dimensional global DINO descriptor.

\subsection{Fusion and regression head.}
We fuse the two branches (2D and 3D branches) by concatenating the 128-dimensional latent code output by the 3D branch with the 256-dimensional DINO embedding into a single 384-dimensional vector. This fused representation passes through a two-layer MLP with sizes 384\(\rightarrow\)256 and 256\(\rightarrow\)128, each with \texttt{LayerNorm}, LeakyReLU ($0.2$), and dropout ($0.3$). The final regression head is a single linear layer mapping the 128-dimensional fused feature to 4 outputs corresponding to the parameters of the Deep Evidential Regression framework: $\gamma$ (the predicted volume or surface area), alongside $v$, $\alpha$, and $\beta$, which are used to explicitly quantify both aleatoric and epistemic uncertainties. To enforce positivity of the predicted geometric attribute ($\gamma$) and ensure valid bounds for the Normal-Inverse-Gamma (NIG) distribution parameters ($v > 0$, $\alpha > 1$, $\beta > 0$), we apply \texttt{softplus} activations appropriately to these outputs.

\subsection{Loss function and uncertainty.}
We train the model using a combined objective consisting of an evidential continuous regression loss~\cite{amini2020deep} and a custom robust volume loss. To capture predictive uncertainty, the network outputs the parameters of a Normal-Inverse-Gamma distribution: the predicted mean $\gamma$, along with parameters $v$, $\alpha$, and $\beta$. The evidential loss is defined as $L_{ev} = L_{NLL} + \lambda L_{R}$, where $L_{NLL}$ is the negative log-likelihood of the target under the predicted distribution, and $L_{R} = |y - \gamma|(2v + \alpha)$ is a regularizer that penalizes high-confidence errors. We set the regularization weight to $\lambda = 0.1$. We additionally apply a robust loss, $L_{rob}$, which combines the Mean Absolute Error (MAE) and the relative error. The robust loss is defined as $L_{rob} = \delta \text{MAE} + (1 - \delta)\text{Relative Error}$. We use a weight of $\delta = 0.7$, leaving a weight of $0.3$ for the relative error.

\subsection{Optimization and training details.}
We optimize all models with AdamW using an initial learning rate of \(1.6\times10^{-4}\) and a weight decay of \(10^{-4}\). We employ a cosine annealing learning-rate schedule (\texttt{CosineAnnealingLR}) with \(T_{\max}=60\) epochs and a minimum learning rate of \(2 \times 10^{-6}\). Training is run for up to 650 epochs; early stopping was applied.
We train with a batch size of 1 (one object per batch), which is sufficient given the large number of points per cloud. We use mixed-precision training via \texttt{torch.cuda.amp} (automatic casting and gradient scaling) to reduce memory usage and accelerate training. Random seeds for PyTorch and NumPy are fixed to 42 (including dataloader workers) to ensure reproducibility.

\subsection{Data augmentation.}
To enhance model robustness and generalization, we apply data augmentation techniques to the datasets during training. This includes random rigid transforms per sample, such as 3D rotations based on independent uniform angles across Euler axes and axis reflections with a $50\%$ probability along each coordinate. These transformations are limited to the \((x,y,z)\) coordinates of the feature vector $\Feat_{3D}$, the confidence scores are left unchanged. Each point cloud undergoes a deterministic preprocessing step before augmentations are applied. We first recenter the 3D coordinates by subtracting their centroid. Next, If a point cloud contains more than a fixed budget of points (\(N_{\max}=50{,}000\)), we randomly subsample \(N_{\max}\) points without replacement; samples with fewer than a minimum number of points (e.g., \(N_{\min}=20{,}000\)) are discarded during dataset construction. 

When Gaussian jitter is enabled, it is implemented as isotropic noise \(\varepsilon \sim \mathcal{N}(0,\sigma^2 I_3)\) on the coordinates with a small standard deviation \(\sigma = 0.001\). All augmentations (rotations, reflections, jitter and subsampling) operate strictly on the \((x,y,z)\) coordinates, leaving features and DINO descriptors unchanged. For both the training and validation sets, images were chosen randomly. In Sec. 5.2 in the main paper, we further consider randomly sampled image subsets with varying numbers of views per object.
At evaluation time, these augmentations are disabled to maintain consistency and ensure reliable, deterministic performance metrics. 

\subsection{Environment.}
The models were run on a single Nvidia A100 GPU on a Linux cluster using Python and PyTorch. Key tools include libraries like PyVista/Trimesh for ground-truth 
computation and rendering, as well as repositories of the pre-trained models utilized for the framework and benchmarks (Trellis~\cite{xiang2025trellis}, MapAnything~\cite{keetha2025mapanything}, VGGT~\cite{wang2025vggt}, 2DGS~\cite{huang20242d}, Agisoft). For training, we used a 80/20 train/validation split.

\section{Ground Truth Calculation}
\label{sec:gt_calculation}
\subsection{Data normalization and volume validation.}
To standardize the scale of the 3D models across our datasets, we evaluated four different spatial normalization techniques. Scale factors were calculated based on the inverse of the maximum dimension of the object's Axis-Aligned Bounding Box (AABB), the Principal Component Analysis Oriented Bounding Box (PCA OBB), the Minimum Oriented Bounding Box (Min OBB), and the diameter of the Minimum Bounding Sphere. While the minimum oriented bounding box (Min OBB) is also rotationally invariant, the bounding sphere strictly guarantees that all input coordinates are normalized within a canonical unit sphere and therefore, it was chosen as the default approach. To ensure these scaling factors were robust to noise, we first applied statistical outlier removal on a random subset of 50,000 vertices, discarding points that deviated by more than three standard deviations from the mean. We computed the ground-truth volume using two parallel strategies to verify accuracy. We utilized a median ray-casting method along the three principal axes on the original non-watertight meshes, while the exact analytical volume and surface area were computed directly on their watertight counterparts. We estimated the signed volume using Trimesh, which sums the volumes of tetrahedra formed by each triangle and the origin. For a triangle with vertices
$\mathbf{v}_{i0}, \mathbf{v}_{i1}, \mathbf{v}_{i2} \in \mathbb{R}^3$,
with a counter-clockwise order when viewed from outside, the formula is provided as follows:
\begin{equation}
V_{\text{signed}}
=
\frac{1}{6}\sum_{i}
\det\!\begin{bmatrix}
\mathbf{v}_{i0} & \mathbf{v}_{i1} & \mathbf{v}_{i2}
\end{bmatrix}
=
\frac{1}{6}\sum_{i}
\mathbf{v}_{i0}\cdot\big(\mathbf{v}_{i1}\times \mathbf{v}_{i2}\big).
\label{eq:signed_volume_formula}
\end{equation}
Surface area, instead, is computed as the exact triangle sum from Trimesh:
 \begin{equation}
 \tilde{A}(\tilde{\mathcal{M}})=
 \sum_{(i,j,k)\in \mathcal{F}}
 \frac{1}{2}\,\big\|(\mathbf{v}_j-\mathbf{v}_i)\times(\mathbf{v}_k-\mathbf{v}_i)\big\|,
 \end{equation}

Finally, to guarantee high-fidelity ground-truth measurements, we strictly filtered the dataset, retaining only the volumes where the discrepancy between the ray-casted (non-watertight) and analytical (watertight) methods was lower than 3\%.

\subsection{Low-Volume Sensitivity and Metric Selection.}
In our evaluation, we report both the Mean and Median Absolute Percentage Error (MAPE and MdAPE) to provide a balanced assessment of reconstruction performance. While the MAPE normalizes the error by the true value, it is highly sensitive to the object's scale. For instance, a small absolute error of \(0.3~\text{cm}^3\) represents only a \(0.33\%\) error for a larger object like an apple of \(90~\text{cm}^3\), but escalates to a \(33\%\) error for a tiny object like a bean of \(0.9~\text{cm}^3\). This disproportionate effect makes the median a valuable complement for capturing central tendencies without being overly influenced by such extremes. Furthermore, factors like minor artifacts in watertight volume calculations can bias the ground-truth volume downward. Because low-volume objects are more susceptible to these inaccuracies, even small geometric artifacts, whether due to watertight processing or ordinary reconstruction noise, can disproportionately inflate the overall mean, underscoring the need for both metrics.

\section{Decoder Selection}
\label{sec:decoder_selection}

To identify the most effective decoder for processing the extracted 3D feature map + confidence scores and predicting volume and surface area, we evaluated several architectures: standard Graph Convolutional Network (GCN)~\cite{kipf2017semi}, Graph Isomorphism Network (GIN)~\cite{xu2019how},  Graph Sample and Aggregate (GraphSAGE)~\cite{Ham2017GraphSAGE}, PointTransformer~\cite{zhao2021point}, PointMLP~\cite{ma2022rethinking}, and Dynamic Graph CNN (DGCNN)~\cite{wang2019dynamic}.

In these experiments, all models were trained on random subsets of $22,000$ points from each point cloud, with a new subset sampled per training epoch. This strategy helps manage GPU memory usage, especially for architectures like DGCNN, whose memory requirements grow with the number of points.

As detailed in Table~\ref{tab:architectures_surf_vol}, DGCNN demonstrated the strongest overall performance, achieving the lowest or second-lowest metrics across most evaluation measures (MAE, MAPE, and RMSE) for both volume and surface area predictions. Consequently, we utilize a DGCNN point-cloud encoder for the experiments in Sections 5.4–5.5 (Tables 4–5). However, for the majority of our experiments that demand significant computing time, we employ GIN. Although GIN ranks as the second-best performing model overall, it is vastly more computationally efficient: processing 50k points on an A100 GPU occupies 38GB of memory with DGCNN, compared to just 3GB with GIN. Thus, GIN represents the most practical choice for most of our evaluations.

\begin{table}[!h]
\centering
\caption{Comparison of different decoder architectures for volume and surface prediction. The \textbf{best} and \underline{second-best} scores are highlighted.}
\label{tab:architectures_surf_vol}
\renewcommand{\arraystretch}{1.2} 
\scriptsize
\begin{tabular}{lcccc}
\toprule
\textbf{Architecture} & \textbf{MAE ↓} & \makecell{\textbf{MAPE ↓}} & \makecell{\textbf{MdAPE ↓} } & \textbf{RMSE ↓} \\
\midrule
GCN              & \underline{1.61e-2} / 8.17e-1          & 22.00 / 14.98          & 13.02 / 11.22          & \underline{2.70e-2} / 1.22 \\
GIN              & 1.81e-2 / \textbf{4.88e-1}             & 25.05 / 13.64          & 14.43 / \underline{9.98} & 2.97e-2 / \underline{1.10} \\
GraphSAGE        & 1.75e-2 / 7.61e-1                      & 22.08 / \textbf{13.07} & 13.55 / 10.22          & 3.01e-2 / 1.18 \\
PointTransformer & 1.71e-2 / 8.04e-1                      & \underline{21.21} / 13.81 & \underline{12.05} / 11.65 & 3.05e-2 / 1.20 \\
PointMLP         & 1.96e-2 / 7.97e-1                      & 28.28 / 14.00          & 14.03 / \textbf{9.45}  & 3.58e-2 / 1.25 \\
DGCNN            & \textbf{1.40e-2} / \underline{6.90e-1} & \textbf{20.23} / \underline{13.57} & \textbf{11.42} / 10.02 & \textbf{2.02e-2} / \textbf{0.95} \\
\bottomrule
\end{tabular}
\end{table}

\section{Additional Results}

Table~\ref{tab:ape_ours_inin_voleta} reports all the results for the CVPR MetaFood Challenge, including our framework with and without deterministic loss, ININ and VolETA~\cite{chen2024metafood3d,almughrabi2024voleta}. 
The table consists of three different types of scenes: simple (first 8 scenes, 200 images) medium (scene between 9 and 15, 30 images) and hard (between 16 and 20, single view).

\begin{table}[t]
\centering
\scriptsize
\setlength{\tabcolsep}{4.5pt}      
\renewcommand{\arraystretch}{1.15} 
\caption{CVPR Metafood Challenge: Per-scene error percentage (APE $=|\hat v - v|/v \cdot 100$). 
Best (lowest) per row in \textbf{bold}. Dashes (--) = not reported.}
\label{tab:ape_ours_inin_voleta}
\begin{tabular}{c c c c c} 
\toprule
\textbf{Idx} & \makecell{\textbf{Ours} \\ \textbf{(Evid.) (\%)}} & \makecell{\textbf{Ours} \\ \textbf{(Evid.+determ.) (\%)}} & \textbf{ININ (\%)} & \textbf{VolETA (\%)} \\
\midrule
1  & 7.22  & 6.75  & 15.52 & \textbf{3.97} \\
2  & \textbf{3.68}  & 5.60  & 14.59 & 22.64 \\
3  & \textbf{0.11}  & 2.59  & 34.62 & 11.69 \\
4  & 14.36 & 20.89 & 17.76 & \textbf{5.46} \\
5  & 2.28  & 3.54  & 0.84  & \textbf{0.81} \\
6  & \textbf{1.28}  & 5.41  & 9.44  & 6.07 \\
7  & 19.45 & 21.84 & 11.86 & \textbf{1.13} \\
8  & 24.71 & 24.05 & \textbf{4.67}  & 7.79 \\
9  & 5.03  & 4.05  & \textbf{0.45}  & 3.72 \\
10 & 17.36 & 10.60 & \textbf{1.86}  & 5.72 \\
11 & 8.13  & 2.06  & \textbf{0.96}  & 5.61 \\
12 & \textbf{2.31}  & 11.78 & --    & --    \\
13 & 30.59 & \textbf{0.83}  & 8.57  & 18.07 \\
14 & 1.70  & \textbf{0.78}  & 12.22 & 9.22 \\
15 & 84.71 & \textbf{63.24} & --    & --    \\
16 & 29.98 & \textbf{28.39} & 32.77 & 37.77 \\
17 & 5.54  & 17.78 & \textbf{0.37}  & 23.56 \\
18 & \textbf{5.97}  & 10.08 & 34.01 & 23.42 \\
19 & 27.19 & 12.29 & 8.64  & \textbf{1.57} \\
20 & \textbf{5.02}  & 7.38  & 20.03 & 9.29 \\
\bottomrule
\end{tabular}
\end{table}

\begin{table*}[t]
\centering
\scriptsize                     
\setlength{\tabcolsep}{4.5pt}   
\renewcommand{\arraystretch}{1.15} 
\caption{Synthetic corals: MAPE and MdAPE for volume and surface (lower is better) for a single image, 5 and 30 input views. The baseline is Agisoft; ``--'' indicates configurations not possible to evaluate for that method.}
\label{tab:corals_single_5_30_trellis_ours_2dgs}
\begin{tabular}{lllcccccc}
\toprule
Dataset & \# Views & Stat. & \multicolumn{2}{c}{\textbf{Ours}} & \multicolumn{2}{c}{\textbf{Trellis}} & \multicolumn{2}{c}{\textbf{Agisoft}} \\
\cmidrule(lr){4-5} \cmidrule(lr){6-7} \cmidrule(lr){8-9}
& & & Vol & Surf & Vol & Surf & Vol & Surf \\
\midrule
\multirow{6}{*}{Corals}
& \multirow{2}{*}{Single image}
& Mean   & \textbf{22.08} & \textbf{13.71} & 87.00 & 64.65 & --    & --    \\
& & Median & \textbf{18.21} & \textbf{10.24} & 85.41 & 67.58 & --    & --    \\
& \multirow{2}{*}{5} 
& Mean   & \textbf{23.36} & \textbf{12.59} & --    & --    & 126.85 & 47.90\\
& & Median & \textbf{15.97} & \textbf{6.28}  & --    & --    & 69.13  & 47.41 \\
& \multirow{2}{*}{30} 
& Mean   & \textbf{19.33} & \textbf{10.28} & --    & --     & 93.12 & 41.49 \\
& & Median & \textbf{15.57} & \textbf{7.53}  & --    & --     & 63.70 & 46.41 \\
\bottomrule
\end{tabular}
\end{table*}

\begin{table}[ht]
\centering
\scriptsize
\setlength{\tabcolsep}{4.5pt}      
\renewcommand{\arraystretch}{1.15} 
\caption{Volume predictions with evidential uncertainty (Evidential Loss Only). Aleatoric and epistemic standard deviations are provided alongside the total standard deviation.}
\label{tab:evid_loss_only}
\begin{tabular}{c c c c c c}
\toprule
\textbf{Idx} & \textbf{Ground Truth} & \textbf{Predicted} & \textbf{Aleatoric STD} & \textbf{Epistemic STD} & \textbf{Total STD} \\
\midrule
1 & 0.1675 & 0.1796 & 0.0237 & 0.0145 & 0.0278 \\
2 & 0.1163 & 0.1206 & 0.0159 & 0.0090 & 0.0183 \\
3 & 0.0235 & 0.0235 & 0.0043 & 0.0020 & 0.0047 \\
4 & 0.0340 & 0.0389 & 0.0069 & 0.0034 & 0.0077 \\
5 & 0.0939 & 0.0918 & 0.0137 & 0.0075 & 0.0156 \\
6 & 0.1067 & 0.1054 & 0.0143 & 0.0080 & 0.0164 \\
7 & 0.1404 & 0.1678 & 0.0206 & 0.0123 & 0.0240 \\
8 & 0.3036 & 0.2286 & 0.0258 & 0.0162 & 0.0305 \\
9 & 0.2834 & 0.2692 & 0.0323 & 0.0210 & 0.0386 \\
10 & 0.0243 & 0.0201 & 0.0044 & 0.0020 & 0.0048 \\
11 & 0.1219 & 0.1318 & 0.0206 & 0.0120 & 0.0238 \\
12 & 0.0675 & 0.0690 & 0.0108 & 0.0057 & 0.0122 \\
13 & 0.0225 & 0.0156 & 0.0035 & 0.0015 & 0.0038 \\
14 & 0.0556 & 0.0566 & 0.0098 & 0.0050 & 0.0111 \\
15 & 0.1099 & 0.2030 & 0.0421 & 0.0267 & 0.0499 \\
16 & 0.2347 & 0.1643 & 0.0218 & 0.0130 & 0.0254 \\
17 & 0.0575 & 0.0607 & 0.0113 & 0.0058 & 0.0127 \\
18 & 0.1068 & 0.1004 & 0.0175 & 0.0098 & 0.0200 \\
19 & 0.0440 & 0.0320 & 0.0069 & 0.0033 & 0.0077 \\
20 & 0.0356 & 0.0338 & 0.0069 & 0.0033 & 0.0077 \\
\bottomrule
\end{tabular}
\end{table}

\begin{table}[ht]
\centering
\scriptsize
\setlength{\tabcolsep}{4.5pt}      
\renewcommand{\arraystretch}{1.15} 
\caption{Volume predictions with evidential uncertainty (Evidential + Deterministic Loss). Aleatoric and epistemic standard deviations are provided alongside the total standard deviation.}
\label{tab:evid_determ_loss}
\begin{tabular}{c c c c c c}
\toprule
\textbf{Idx} & \textbf{Ground Truth} & \textbf{Predicted} & \textbf{Aleatoric STD} & \textbf{Epistemic STD} & \textbf{Total STD} \\
\midrule
1 & 0.1675 & 0.1788 & 0.0225 & 0.0139 & 0.0265 \\
2 & 0.1163 & 0.1228 & 0.0158 & 0.0092 & 0.0183 \\
3 & 0.0235 & 0.0229 & 0.0049 & 0.0024 & 0.0054 \\
4 & 0.0340 & 0.0411 & 0.0070 & 0.0036 & 0.0079 \\
5 & 0.0939 & 0.0973 & 0.0142 & 0.0081 & 0.0163 \\
6 & 0.1067 & 0.1010 & 0.0140 & 0.0080 & 0.0161 \\
7 & 0.1404 & 0.1711 & 0.0195 & 0.0119 & 0.0229 \\
8 & 0.3036 & 0.2306 & 0.0240 & 0.0153 & 0.0285 \\
9 & 0.2834 & 0.2719 & 0.0308 & 0.0201 & 0.0367 \\
10 & 0.0243 & 0.0269 & 0.0055 & 0.0027 & 0.0062 \\
11 & 0.1219 & 0.1244 & 0.0188 & 0.0111 & 0.0219 \\
12 & 0.0675 & 0.0754 & 0.0117 & 0.0064 & 0.0134 \\
13 & 0.0225 & 0.0227 & 0.0048 & 0.0023 & 0.0053 \\
14 & 0.0556 & 0.0552 & 0.0091 & 0.0048 & 0.0103 \\
15 & 0.1099 & 0.1794 & 0.0321 & 0.0201 & 0.0379 \\
16 & 0.2347 & 0.1681 & 0.0200 & 0.0122 & 0.0234 \\
17 & 0.0575 & 0.0677 & 0.0110 & 0.0059 & 0.0125 \\
18 & 0.1068 & 0.0960 & 0.0148 & 0.0084 & 0.0171 \\
19 & 0.0440 & 0.0386 & 0.0075 & 0.0038 & 0.0084 \\
20 & 0.0356 & 0.0382 & 0.0069 & 0.0035 & 0.0077 \\
\bottomrule
\end{tabular}
\end{table}

Table~\ref{tab:corals_single_5_30_trellis_ours_2dgs} compares our framework, with a DGCNN decoder, on synthetic corals against an MVS-based baseline using Agisoft and the Trellis pipeline. We evaluate three input regimes: single view, 5 views, and 30 views on a subset of 15 synthetic coral assets drawn from the main paper test set. For the single-view setting, we render 35 independent views per asset and run inference on each view separately; we then aggregate the resulting volume and surface predictions across all views and assets to report MAPE and MdAPE.

\begin{figure*}[t]
  \centering
  \begin{subfigure}{0.39\textwidth}
    \centering
    \includegraphics[width=\linewidth]{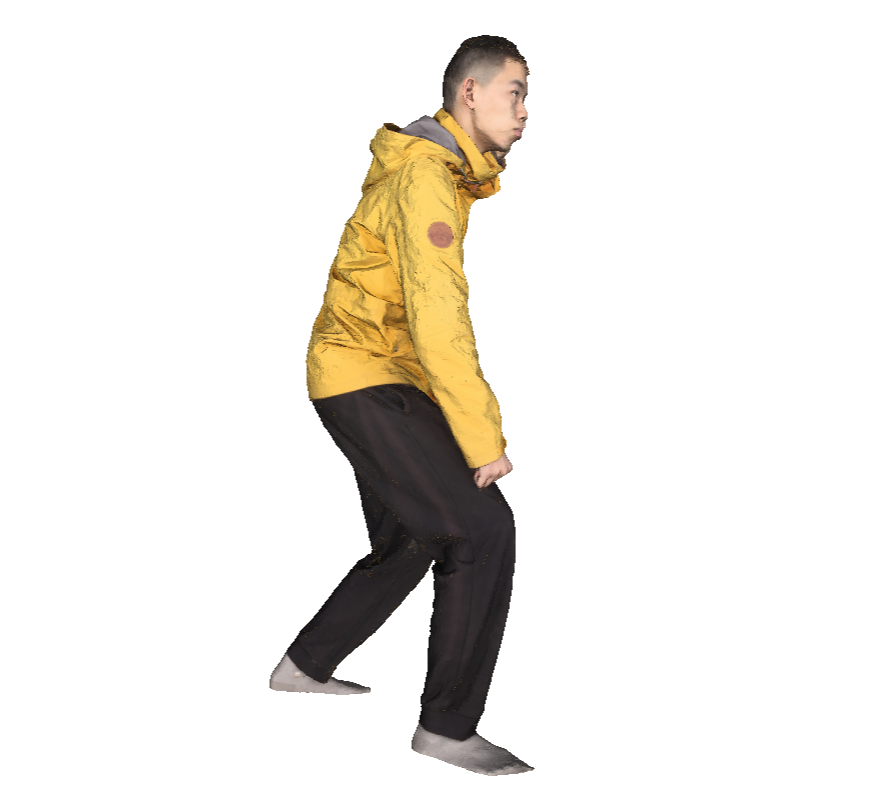}
    \caption{}
    \label{fig:human_gt04}
  \end{subfigure}
\hspace{0.025\textwidth}
  \begin{subfigure}{0.39\textwidth}
    \centering
    \includegraphics[width=\linewidth]{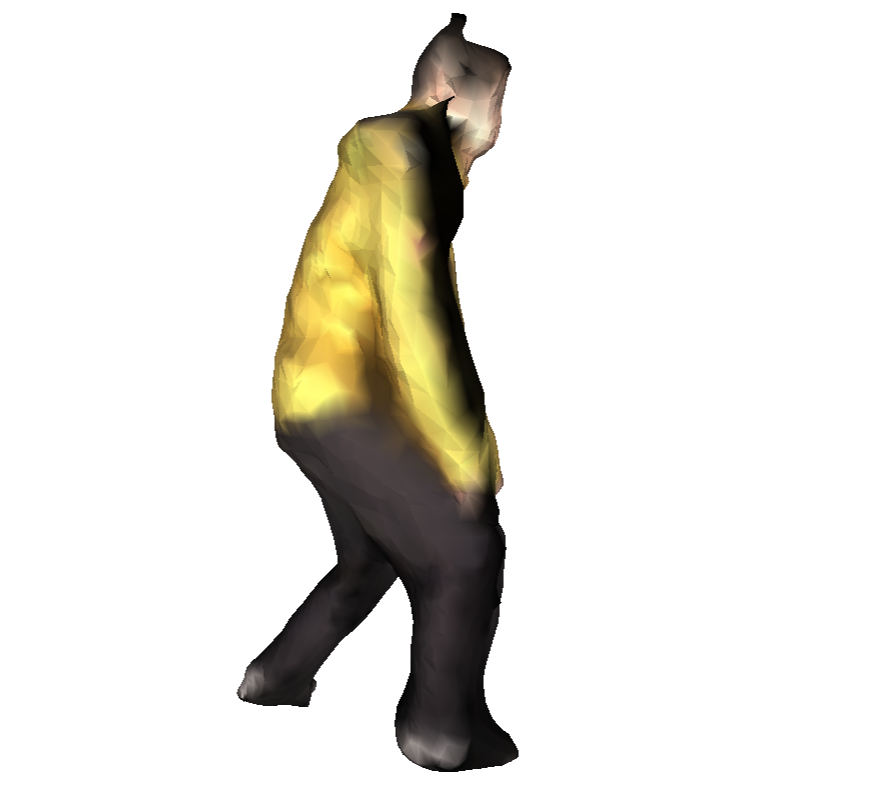}
    \caption{}
    \label{fig:human_compare04}
  \end{subfigure}
  \caption{Left: ground-truth image. Right: Agisoft reconstruction using 30 images, which fails to recover the correct geometry and appearance in most cases.}
  \label{fig:human_side_by_side_04}
\end{figure*}

For the case of 5 and 30 images, we first use Agisoft to reconstruct a mesh from the multi-view images, then we postprocess the mesh to remove some edge vertices that represent the black background and convert it to a watertight mesh. Finally, we use Trimesh to compute the volume and surface as described in the main paper. 
This shows that our learned regressor can extract reliable 3D cues even in the most challenging single-view setting, where Agisoft cannot be applied. When more views are available, our performance remains strong or improves slightly. Furthermore, it vastly outperforms Trellis on a single image which also fails to accurately regressing volume and surface area from the reconstructed meshes.
Overall, our approach is both more data efficient, performing well even from a single image, and more robust as the number of views increases; while Agisoft and Trellis either cannot be evaluated in some settings or remain significantly less accurate (see Fig.~\ref{fig:human_side_by_side_04}).

\paragraph{Uncertainty.}
Tables~\ref{tab:evid_loss_only} and~\ref{tab:evid_determ_loss} report the per-sample ground truth, predicted volumes, and estimated uncertainties for the 20 evaluation samples. To ensure the uncertainty metrics remain interpretable and share the same physical units as the volume predictions, we report them as standard deviations rather than raw variances. Specifically, the tables detail the decomposed aleatoric standard deviation (representing inherent data noise) and epistemic standard deviation (representing model uncertainty). The total standard deviation is also provided, calculated as the square root of the total predictive variance (the sum of the aleatoric and epistemic variances). Comparing the two tables illustrates the per-sample behavior of the framework when optimized with the evidential loss alone versus the combined evidential and deterministic objective.

\end{document}